\documentclass[compsocconference]{IEEEtran}
\usepackage{cite}
\ifCLASSINFOpdf
\else
   \usepackage[dvips]{graphicx}
   \DeclareGraphicsExtensions{.eps}
\fi
\usepackage{amssymb}
\usepackage{bbm}
\usepackage[cmex10]{amsmath}
\usepackage{algorithm}
\usepackage{algorithmic}
\usepackage{array}
\usepackage{mdwmath}
\usepackage{mdwtab}
\usepackage[tight,footnotesize]{subfigure}

\usepackage{stfloats}
\usepackage{url}
\usepackage{bbm}
\usepackage{multirow}
\usepackage{color}

\begin{document}
\title{Discriminative Clustering with Relative Constraints}

\author{\IEEEauthorblockN{Yuanli Pei\IEEEauthorrefmark{1},
Xiaoli Z. Fern\IEEEauthorrefmark{1},
R\'omer Rosales\IEEEauthorrefmark{2}, and
Teresa Vania Tjahja\IEEEauthorrefmark{1}
}

\IEEEauthorblockA{\IEEEauthorrefmark{1}School of EECS,
Oregon State University.\\ Email: \{peiy, xfern, tjahjat\}@eecs.oregonstate.edu }

\IEEEauthorblockA{\IEEEauthorrefmark{2}LinkedIn.
Email: rrosales@linkedin.com}
}


\maketitle

\begin{abstract}
We study the problem of clustering with \emph{relative constraints}, where each
constraint specifies relative similarities among instances.
In particular, each constraint $(x_i, x_j, x_k)$ is acquired by posing a query:
\emph{is instance $x_i$ more similar to $x_j$ than to $x_k$?}
We consider the scenario where answers to such queries are based on an underlying (but unknown)
\emph{class concept}, which we aim to discover via clustering.
Different from most existing methods that only consider constraints derived
from \emph{yes} and \emph{no} answers, we also incorporate \emph{don't know}
responses. We introduce a \textbf{D}iscriminative \textbf{C}lustering method
with \textbf{R}elative \textbf{C}onstraints (DCRC) which assumes a
natural probabilistic relationship between instances, their underlying
cluster memberships, and the observed constraints. The objective is to maximize
the model likelihood given the constraints, and in the meantime
enforce cluster separation and cluster balance by also
making use of the unlabeled instances.
We evaluated the proposed method using constraints generated from ground-truth class labels,
and from (noisy) human judgments from a user study.
Experimental results demonstrate: 1) the usefulness of relative constraints,
in particular when \emph{don't know} answers are considered; % in the model;
2) the improved performance of the proposed method over state-of-the-art methods that
utilize either relative or pairwise constraints; and 3) the robustness of our method in
the presence of noisy constraints, such as those provided by human judgement.

\end{abstract}



% \fi
\IEEEpeerreviewmaketitle

\section{Introduction}
\label{sec:intro}
Unsupervised clustering can be improved with the aid of side information
%, exposing instance-level relationships
for the task at hand. In general, side information refers to knowledge
beyond instances themselves that can help inferring the underlying instance-to-cluster assignments.
%, %instance-to-cluster assignments,
One common and useful type of side information has been
represented in the form of  %(possibly soft)
\emph{instance-level constraints} that expose instance-level relationships.
%In this paper, we focus on such type of information.

Previous work has primarily focused on the use of
\emph{pairwise constraints} (e.g.,\cite{wagstaff01,xing2003NIPS,basu04,
shental03NIPS,lu04NIPS,basu04KDD,lange05CVPR,nelson07ICML,ge2007KDD,
lu07JMLR, basu2008Book}), where a pair of instances is indicated to
belong to the same cluster by a Must-Link (ML) constraint or to
different clusters by a Cannot-Link (CL) constraint.
%The readers are reffered to \cite{basu2008Book} for a comprehensive introduction.
More recently, various studies \cite{schultz2003NIPS,rosales2006KDD, huang2011, liu2012ICDM,kumar08TKDE,liu2011KDD} have suggested that
domain knowledge can also be incorporated in the form of relative comparisons or \emph{relative
constraints}, where each constraint specifies {\it whether instance $x_i$ is more similar to $x_j$ than to $x_k$}.
\begin{figure*}[ht!]
\centering
\begin{tabular}{ccccccccccc}
        \hspace{0.2cm}
        \includegraphics[width=0.022\textwidth]{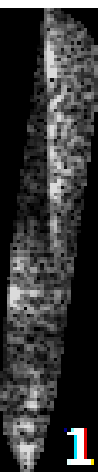}  & %\hspace{0.2cm}
        \includegraphics[width=0.022\textwidth]{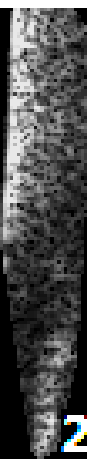}  & \hspace{1.5cm}
        \includegraphics[width=0.022\textwidth]{figure/syllable1.eps}  & %\hspace{0.25cm}
        \includegraphics[width=0.022\textwidth]{figure/syllable2.eps}  & %\hspace{0.05cm}
        \includegraphics[width=0.022\textwidth]{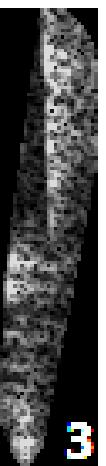}  & \hspace{1.5cm}
        \includegraphics[width=0.022\textwidth]{figure/syllable1.eps}  & %\hspace{0.25cm}
        \includegraphics[width=0.023\textwidth]{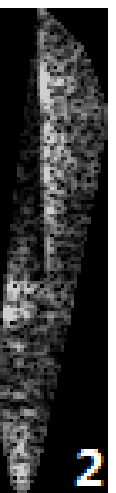}  & %\hspace{0.05cm}
        \includegraphics[width=0.022\textwidth]{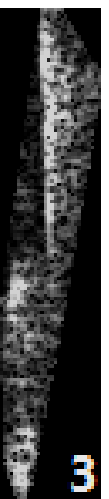}  & \hspace{1.5cm}
        \includegraphics[width=0.022\textwidth]{figure/syllable1.eps}  & %\hspace{0.05cm}
        \includegraphics[width=0.055\textwidth]{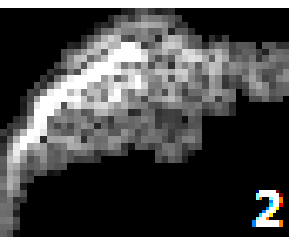}   & %\hspace{0.05cm}
        \includegraphics[width=0.13\textwidth]{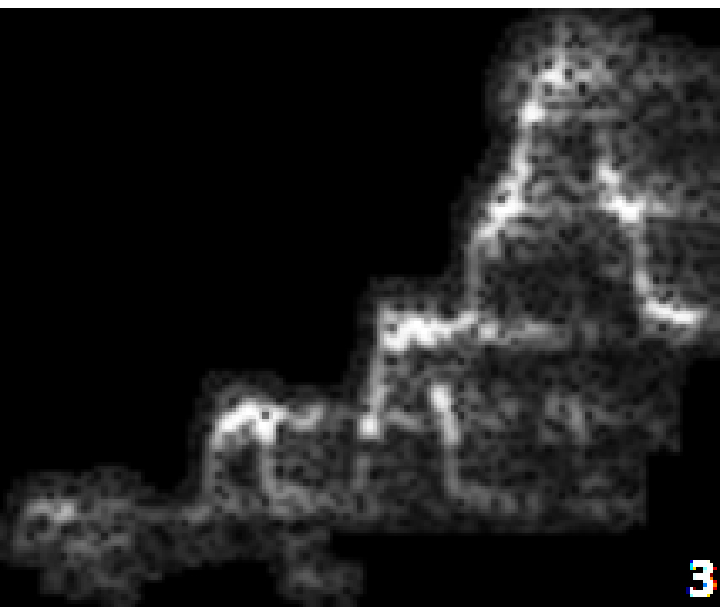}     \\
        \multicolumn{2}{c}{\footnotesize{(a) Pairwise Const.}}&
        \multicolumn{3}{c}{\hspace{1.5cm}\footnotesize{(b) Relative Const.}}&
        \multicolumn{3}{c}{\hspace{1.5cm}\footnotesize{(c) Relative Const.}} &
        \multicolumn{3}{c}{\hspace{1.5cm}\footnotesize{(d) Relative Const.}}\\
\end{tabular}
\caption{\label{fig:example} Examples for labeling pairwise vs.\ relative constraints from {\em Birdsong} data.
Labeling question for (a): \emph{Do syllable 1 and syllable 2 belong to the same cluster?} Labeling question for (b) (c) and (d):
\emph{Is syllable 1 more similar to syllable 2 than to syllable 3?}
(a) and (b) reveal the cases where relative constraints are easier to label. The cases in (c) and (d) motivate the introducing
of a \emph{don't know} answer for relative constraints. }
\end{figure*}%%

%\begin{figure}[b]
%\centering
%     \subfigure[Pairwise: Do syllable 1 and syllable 2 belong to the same cluster?]{ \label{fig:example_pair}
%            \begin{tabular}{cc}
%                \includegraphics[width=0.05\textwidth]{figure/temp3_39_1_new.eps}  & \hspace{0.2cm}
%                \includegraphics[width=0.05\textwidth]{figure/temp4_70_1_new.eps}
%             \end{tabular}
%            }
%     \hspace{0.65cm}
%     \subfigure[Relative: Is syllable 1 more similar to syllable 2 than to syllable 3?]{ \label{fig:example_triple}
%            \begin{tabular}{ccc}
%                    \includegraphics[width=0.05\textwidth]{figure/temp3_39_1_new.eps}  & \hspace{0.25cm}
%                    \includegraphics[width=0.05\textwidth]{figure/temp4_70_1_new.eps}  & \hspace{0.05cm}
%                    \includegraphics[width=0.05\textwidth]{figure/temp3_69_2_new.eps}
%            \end{tabular}
%                    }
%\caption{\label{fig:example} Example questions for labeling pairwise/relative constraints from {\em Birdsong} data.}
%\end{figure}
We were motivated to focus on relative constraints for a couple of reasons.
First, the labeling (proper identification) of relative constraints by humans
appears to be more reliable than that of pairwise constraints. Research in psychology
has revealed that people are often inaccurate in making absolute judgments (required for pairwise constraints),
but they are more trustworthy when judging comparatively\cite{nunnally94}.
Consider one of our applications, where we would like to form clusters
of bird song syllables based on spectrogram segments from recorded
sounds. Figure 1(a) and 1(b) shows examples of the two types of
constraints/questions considered.
In the examples, syllable $1$ in both figures and
syllable $3$ in 1(b) are from the same singing pattern and syllable $2$ in both figures belongs to a
different one.
From the figures, it is apparent that making an
absolute judgment for the pairwise constraint in 1(a)
is more difficult. In contrast, the comparative question for labeling relative
constraint in 1(b) is much easier to answer.
Second, since each relative constraint includes information about three instances,
they tend to be more informative than pairwise constraints (even when several pairwise
constraints are considered). This is formally characterized in Section \ref{sec:info}.
%relative constraints are generally more informative than pairwise constraints.
%Therefore, our work focuses on the form of relative constraints.

In the area of learning from relative constraints, most work uses
metric learning approaches \cite{schultz2003NIPS,rosales2006KDD,huang2011, liu2012ICDM,kumar08TKDE}.
Such approaches assume that there is an underlying metric that
determines the outcome of the similarity comparisons,
%, i.e., $x_i$ is
%believed to be more similar to $x_j$ than to $x_k$ if $d(x_i, x_j) <
%d(x_i, x_k)$, where $d(\cdot)$ is the underlying distance function.
and the goal is to learn such a metric. The learned metric is often later used
for clustering ({\it e.g.}, via Kmeans or related approaches).
%\cite{schultz2003NIPS,rosales2006KDD,huang2011, liu2012ICDM}. %, or at the same time find a clustering solution \cite{kumar08TKDE}.
In practice, however, we may not have access to an \emph{oracle} metric. Often the constraints are provided
%based on some underlying \emph{class concept}
in a way that instances from the same class are considered more similar than those from different classes.
This paper explicitly considers such scenarios where constraints are provided based on
the underlying  \emph{class concept}. Unlike the metric-based
approaches, we aim to directly infer an optimal clustering of the data using the
provided relative comparisons, without requiring explicit metric learning.
%For instance, in the example shown in
%Figure \ref{fig:example_triple}, the three syllables belong to two different
%singing patterns (classes), thus the answer to the question is {\it no}.
%In these scenarios, we seek to infer an optimal clustering of the data using the
%provided relative comparisons as indirect observations or evidence of
%the underlying cluster partitions.

%This work considers the problem of clustering with relative constraints.
Formally, we regard each constraint $(x_i, x_j, x_k)$ as being obtained by
asking: \emph{is $x_i$ more similar to $x_j$ than to $x_k$}, and the answer is
provided by a user/oracle based on the underlying instance clusters. In particular,
%with some probability
a \emph{yes} answer is given if $x_i$ and $x_j$
are believed to belong to the same cluster while $x_k$ is believed to be from a different one.
Similarly, the answer will be \emph{no} if it is believed  that $x_i$ and $x_k$ are in the
same cluster which is different from the one containing $x_j$.
Note that for some triplets, it may not be possible to provide a \emph{yes} or \emph{no} answer.
For example, if the three instances belong to the same cluster, as shown in figure 1(c);
or if each of them  belongs to a different cluster, as shown in figure 1(d).
%However, a user may not be able to provide a \emph{yes} or \emph{no} answer in some cases, which
Such cases have been largely ignored by prior studies. Here, we allow the user to provide a \emph{don't
know} answer (\emph{dnk}) when \emph{yes}/\emph{no} can not be determined.
Such \emph{dnk}'s not only allow for improved labeling flexibility, but also provide
useful information about instance clusters that can help improve clustering,
as will be demonstrated in Section \ref{sec:info} and the experiments.
%(as demonstrated in Section \ref{sec:info} \& Section \ref{sec:experiments}).
%and show how it is possible to
%gain some information from it (Section \ref{sec:info} \& Section \ref{sec:experiments}).

In this work, we introduce a discriminative clustering method, DCRC, that learns
from relative  constraints with \emph{yes}, \emph{no}, or \emph{dnk} labels (Section \ref{sec:model}).
DCRC uses a probabilistic model that naturally connects the instances, their underlying
cluster memberships, and the observed constraints.
%In particular, we use a multi-class logistic model for $P(y|x)$, where $y$ is the cluster label and $x$ is the observed instance.
%Furthermore, we introduce a discrete distribution capturing the probability of observing a constraint label
%based on the instance cluster memberships.
Based on this model, we present a maximum-likelihood objective with additional
terms enforcing cluster separation and cluster balance.
Variational EM is used to find approximate solutions (Section \ref{sec:optimization}).
In the experiments (Section \ref{sec:experiments}), we first evaluate our method on both UCI
and additional real-world datasets with simulated noise-free constraints generated from ground-truth class labels.
The results demonstrate the usefulness of relative constraints including \emph{don't know} answers, and
the performance advantage of our method over current state-of-the-art methods for both relative and pairwise constraints.
We also evaluate our method with human-labeled noisy constraints collected from a user study, and results
show the superiority of our method over existing methods in terms of
robustness to the noisy constraints.

%\section{Constraint Analysis and Problem Statement}
%\section{PROBLEM STATEMENT AND ANALYSIS}
\section{Problem Analysis}
\label{sec:problem}
%\input{problem.tex}
%\subsection{Information: Relative vs. Pairwise Constraints}
%\vspace{-.1in}
%This section analyzes and formally states the problem.
In this section, we first compare the cluster label information obtained
by querying different types of constraints, analyzing the usefulness of relative constraints.
Then we formally state the problem.
%\vspace{-.3in}
\subsection{Information from Constraints}
\label{sec:info}
%\vspace{-.1in}
%\begin{table}
% \caption{\label{tab:triple_pair}Assignment of constraint labels}
%\begin{center}
% \subtable[Relative Const.]{
%    \centering
%    \renewcommand{\arraystretch}{1.1}
%    \label{tab:triple}
%    \begin{tabular}{c|c}
%       {\bf CASES}               &   $l_t$     \\
%       \hline \\
%      $y_{t_1}= y_{t_2}, ~ y_{t_1}\neq y_{t_3}$    & \emph{yes} \\
%      $y_{t_1}= y_{t_3}, ~ y_{t_1}\neq y_{t_2}$    & \emph{no}  \\
%            o.w. & \emph{dnk}
%    \end{tabular}
%  }
%  \subtable[Pairwise Const.]{
%    \centering
%    \renewcommand{\arraystretch}{1.45}
%    \label{tab:pair}
%    \begin{tabular}{c|c}
%        {\bf  Cases }            &  $l'_b$  \\
%          \hline \\
%      $y_{b_1} = y_{b_2}$    & ML        \\
%      $y_{b_1} \neq y_{b_2}$ & CL     \\
%    \end{tabular}
%  }
%\end{center}
%\end{table}
Here we provide a qualitative analysis with a simplified but illustrative example.
Suppose we have $N$  i.i.d instances $\{x_i\}_{i=1}^N$ sampled from $K$ clusters
with even prior $1/K$.  Consider a triplet $(x_{t_1}, x_{t_2}, x_{t_3})$
and a pair $(x_{b_1}, x_{b_2})$. Let  $Y_t = [y_{t_1}, y_{t_2}, y_{t_3}]^T$
and  $Y_b = [y_{b_1}, y_{b_2}]^T$  be their corresponding cluster labels.
Let $l_t \in \{\text{\emph{yes}},\text{\emph{no}}, \text{\emph{dnk}}\}$ and
$l'_b \in \{\text{ML}, \text{CL}\}$ be the label for the relative and pairwise constraint
respectively. In this example they are determined by
\begin{equation}
   \label{eq:triple}
l_t = \left\{
    \begin{array}{c c}
       \emph{yes}, & \text{if} ~~ y_{t_1}= y_{t_2}, ~ y_{t_1}\neq y_{t_3}     \\
       \emph{no} , & \text{if} ~~ y_{t_1}= y_{t_3}, ~ y_{t_1}\neq y_{t_2}     \\
       \emph{dnk}, &   o.w.
    \end{array}
            \right.
\end{equation}
\begin{equation}
    \label{eq:pair}
l'_b = \left\{
    \begin{array}{c c}
        ML, &  \text{if} ~~ y_{b_1} = y_{b_2}   \\
        CL, &  \text{if} ~~ y_{b_1} \neq y_{b_2} ~.
    \end{array}
        \right.
\end{equation}
We can derive the mutual information between a relative constraint and the associated instance
cluster labels as  (see Appendix for the derivation)
\begin{equation}
 \label{eq:i_triple}
 \begin{array} {rl}
 I(Y_t;l_t)    = &  \hspace{-0.2cm}   2\log K - (1-P_{\text{\emph{dnk}}} )\log (K-1) \\
                 &  \hspace{-0.2cm} - P_{\text{\emph{dnk}}}  \log [K^2-2(K-1)] ,
\end{array}
\end{equation}
and that for a pairwise constraint as
\begin{equation}
 \label{eq:i_pair}
 \begin{array} {r l}
     I(Y_b;l'_{b})
  = &\hspace{-0.2cm}\displaystyle  \log K -   P_{CL}  \log (K-1),
\end{array}
\end{equation}
where $P_{\text{\emph{dnk}}} = 1- {2(K-1)}/{K^2}$, and $P_{CL}  = 1- {1}/{K}$.

\begin{figure}
\centering
\includegraphics[width = .45\textwidth]{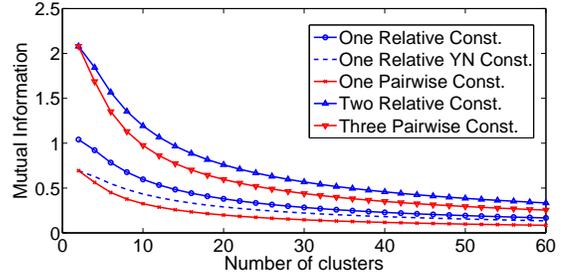}
\caption{Mutual information between instance cluster labels and constraint labels as a
function of the number of clusters.
    %The mutual information of two pairwise constraints and three relative constraints are simply
    %the corresponding times of the value for that of one constraint.
    }
\label{fig:plotMI}
\end{figure}
Figure \ref{fig:plotMI} plots the values of (\ref{eq:i_triple}) and
(\ref{eq:i_pair}) as a function of the number of clusters $K$. Comparing the values of
\emph{one relative const} and \emph{one pairwise const}, we see that, in the
absence of other information, a relative constraint provides more
information. One might argue that labeling a triplet requires inspecting more instances than labeling
a pair, making this comparison unfair. To address this bias,
%suppose that the effort of labeling each constraint is proportional to the number of  involved instances,
%\footnote{Arguably,
%  the effort of labeling the desired amount of relative constraints is
%  likely smaller than labeling the necessary amount of pairwise
%  constraints, since it is usually easier to make a comparative
%  judgment for relative constraints than an absolute judgment for
%  pairwise constraints as discussed in Section \ref{sec:intro}.},
we compare the information gain from the two types of constraints
with the same number of instances, namely, comparing the values
of {\it two relative constraints} with that of {\it three pairwise constraints},
both involving six instances. In Figure \ref{fig:plotMI} we see again that
relative constraints are more informative.
%Such observation is consistent with the empirical results in previous work \cite{rosales2006KDD,kumar08TKDE}
%that clustering methods with relative constraints outperform those with pairwise constraints.

Another aspect worth evaluating is the motivation behind explicitly using {\it dnk} constraints.
In prior work on learning from relative constraints, the constraints are typically
generated by randomly selecting triplets and producing constraints based their class labels.
If a triplet can not be definitely labeled with \emph{yes} or \emph{no}, the resulting constraint is not
employed by the learning algorithm
% Existing metric learning approaches implicitly assume that the underlying metric should respect the class (clusters) labels.
%% \cite{schultz2003NIPS,rosales2006KDD,kumar08TKDE, liu2012ICDM}.
%As shown in their experimental evaluation, instance class labels are used to generate
%constraints. When a \emph{yes} or \emph{no} answer can not be assigned,
%the constraint is not employed by the learning algorithm
(it is ignored). Such methods are by construction not using the information provided by {\it dnk} answers.
%In a realistic scenario, a human labeler providing answers about such relative
%relationships may be uncertain, and may not be able to provide neither
%a \emph{yes} nor a \emph{no} answer. Thus, our interest in using and trying to extract the most
%information from {\it dnk} answers.
%%Yuanli: Removed this. I think here we don't need to emphasize the dnk cases in realistic cases,
%%         since it is already covered in the intro. Adding the discussion here makes the paragraph a bit out of focus.
However, it is possible to show that in general {\it dnk}'s can provide information
about instance labels. If {\it dnk}'s are ignored, the mutual information
can be computed by replacing $H(Y_t|l_t=\text{\emph{dnk}})$ with $H(Y_t)$,
%in the derivation of Eq.\ (\ref{eq:i_triple}),
meaning that the \emph{dnk}'s are not informative about the instance labels. In this case, we have
\begin{equation}
\label{eq:i_triple_yn}
%\nonumber
  I'(Y_t;l_t)  =  2(1-P_{\text{\emph{dnk}}})\log K - (1-P_{\text{\emph{dnk}}} )\log (K-1).
\end{equation}
Comparing the values of \emph{one relative YN const } (which ignores
{\it dnk}) with that of \emph{one relative const} in Figure
\ref{fig:plotMI}, we see a clear gap between using and not using
\emph{dnk} constraints, implying the informativeness of \emph{dnk} constraints.
Additionally, the amount of \emph{dnk} constraints is usually large,
%in some applications,
especially when the number of clusters is large.
Consider randomly selecting triplets from clusters with equal sizes.
There is a $50\%$ chance of acquiring \emph{dnk} constraints in two-cluster problems,
and the chance increases to $78\%$ in eight-cluster problems.
The information provided by such large amount of \emph{dnk} constraints is
substantial.  Hence, we believe it will be beneficial to explicitly
employ and model \emph{dnk} constraints.

\subsection{Problem Statement}
\label{sec:notation}
Let $X = [x_1, \ldots, x_N]^T$ be the given data, where each $x_i \in \mathcal{R}^d$
and $d$ is the feature dimension. Let $Y=[y_1, \ldots, y_N]^T$ be the hidden cluster
label vector, where $y_{i}$ is the label of $x_{i}$. With slight abuse of notation,
we use $\{(t_1, t_2, t_3)\}_{t=1}^M$ to denote the  index set of $M$ triplets,
representing $M$ relative constraints. Each $(t_1, t_2, t_3)$
contains the indices for the three instances in the $t$-{th} constraint.
Let $L = [l_1, \ldots, l_M]^T$ be the constraint label vector,
where $l_t \in \{\text{\emph{yes}, \emph{no}, \emph{dnk}}\}$
is the label of $(x_{t_1},x_{t_2},x_{t_3})$.
Each $l_t$ specifies the answer to the question: \emph{is $x_{t_1}$ more similar to $x_{t_2}$ than to $x_{t_3}$?}
% Romer Removed because we are not making such as strong assumption
%%%%% ,and the value of $l_t$ is given based on Table \ref{tab:triple}.
Our goal is to \emph{partition  the data into $K$ clusters
such that similar instances are assigned to the same cluster, while respecting the given constraints}.
% Romer note: note we cannot say we satisfy all the constraints,
In this paper, we assume that $K$ is pre-specified.

In the following, we will use $I_{t} = \{t_1, t_2, t_3\}$ to denote the set of indices in
the $t$-{th} triplet, use $I$ to index all the distinct instances involved in the constraints, i.e.,
%\begin{equation}
$\label{eq:index_const}
   I = \left\{ 1 \leq i \leq N : i \in \cup_{t=1}^{M} I_t \right\}$,
%\end{equation}
and use $U$ to index the instances that are not in any constraints.

\section{Methodology}
\label{sec:model}
In this section, we introduce our probabilistic model and present the proposed objective functions based on this model.
\subsection{The Probabilistic Model}
\label{sec:model_assumption}
\begin{figure}
\centering
\includegraphics[width = .3\textwidth]{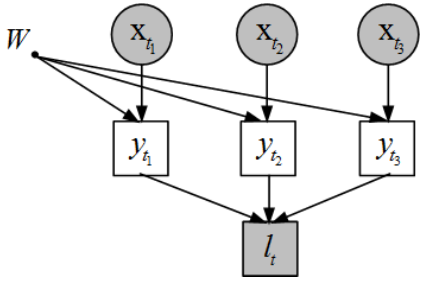}
\caption{The dependencies between three instances $(x_{t_1}, x_{t_2}, x_{t_3})$,
    their cluster labels $(y_{t_1}, y_{t_2}, y_{t_3})$, and the constraint label $l_t$.
}
\label{fig:model}
\end{figure}
We propose a \emph{Discriminative Clustering} model for
\emph{Relative Constraints} (DCRC). Figure \ref{fig:model} shows the proposed probabilistic model defining the dependencies between the input instances $(x_{t_1}, x_{t_2},
x_{t_3})$, their cluster labels $(y_{t_1}, y_{t_2}, y_{t_3})$, and the
constraint label $l_t$ for only one relative constraint. For a collection of
constraints, it is possible to have $y$ variables connected to more than one (or none) constraint label $l$ if
some instances appear in multiple constraints (or do not appear in any
given constraint). %Similarly a single $x$ may be connected with zero,
%one, or more variables $y$.

We use a multi-class logistic classifier to model the conditional
probability of $y$'s given the observed $x$'s.
%(Note that we can easily extend this linear model to handle nonlinear
%cluster structures by considering appropriate kernel functions.)
%Yuanli Comments: I think mention the kernel function here is distracting. To me,
%it seems more natural to mention it later after we explained all the derivations about
%the linear model.
For simplicity, in the following we will use the same notation $x$ to
represent the $(d+1)$-dimensional augmented vector $[x^T, 1]^T$. Let
$W = [w_1, \ldots, w_K]^T$ be a weight matrix in $\mathcal{R}^{K
  \times (d+1)}$, where each $w_k$ contains weights on the $d$-dimensional
feature space and an additional bias term.  Then the conditional probability
is represented as
\begin{equation}
\label{eq:LR}
    P(y=k|x; W) = \frac{\exp{(w_k^T x})}{\sum_{k'}\exp{(w_{k'}^T x})} ~.
\end{equation}

%
%\begin{figure}
%\centering
%\includegraphics[width = .43\textwidth]{figure/graph}
%\caption{ LEFT: The graphical model represents the relationship between the observed features,
%    instance cluster labels, and the constraint label for triplet  $(x_{t_1}, x_{t_2}, x_{t_3})$.
%    RIGHT: An illustrative example describes the model for unconstrained instances $x_n$ and
%    instances that appear in multiple constraints, i.e., $x_j$ and $x_k$ appear in
%    both $(x_i, x_j, x_k)$ and $(x_j, x_k, x_m)$.
%%    For simplicity, we only  show such relationships for one constraint. Note that there might be multiple constraints and it is possible to have
%%    $y$'s  connecting with more than one $l_t$'s if some instances appear in multiple constraints.
%}
%\label{fig:model}
%\end{figure}

In our model, the observed constraint labels only depend on the
cluster labels of the associated instances. In an ideal scenario, the
conditional distribution of $l_t$ given the cluster labels would be
deterministic, as described by Eq.\ (\ref{eq:triple}).  However, in
practice users can make mistakes and %, in general,
be inconsistent during the annotation process. We address this by relaxing
%we may encounter annotation errors. We address this by relaxing
%introducing a parameter $\epsilon$ that relaxes
the deterministic relationship to the distribution $P(l_t|Y_t)$ described in Table \ref{tab:label_dist}.
The relaxation is parameterized by $\epsilon \in [0,1)$,
indicating the probability of an error when answering the query.
Here we let the two erroneous answers have equal probability $\epsilon/2$.
Namely, the ideal label of $l_t$ (e.g., $l_t =\text{\emph{yes}}$ if $y_{t_1}=y_{t_2}, y_{t_1}\neq
y_{t_3}$) is given with probability $1-\epsilon$, and any other labels
(\emph{no} and \emph{dnk} in this case) are given with equal
probability $\epsilon/2$.  In practice, lower values of
$\epsilon$ are expected when constraints have fewer noise. %are obtained from experienced labelers.
Alternatively, we can view this relaxation as allowing the constraints to be
{\it soft} as needed, balancing the trade-off between finding
large separation margins among clusters and satisfying all the constraints.
%It is worth noting that when $\epsilon = 2/3$, the constraint labels are regarded as uniformly distributed
%regardless of the cluster labels. In this case, our model
%naturally reduces to learning without constraints as will be discussed in Section~\ref{sec:estep}.
%Yuanli: Removed this since it is mentioned in section \ref{sec:estep}.
\begin{table}
\caption{\label{tab:label_dist}Distribution of $P(l_t|Y_t)$, $Y_t= [y_{t_1},y_{t_2},y_{t_3}]$. }
\begin{center}
\begin{tabular} {|c|ccc|}
\hline
 Cases & $l_t = \text{\emph{yes}}$  &  $l_t = \text{\emph{no}}$   &$l_t = \text{\emph{dnk}}$      \\ \hline  \hline
$y_{t_1} = y_{t_2}, y_{t_1} \neq y_{t_3} $& $1 - \epsilon$  & $\epsilon/2$   & $\epsilon/2$ \\% \hline
$y_{t_1} = y_{t_3}, y_{t_1} \neq y_{t_2} $& $\epsilon/2$  & $1 - \epsilon$  & $\epsilon/2$  \\% \hline
    o.w.                                    &   $\epsilon/2$   & $\epsilon/2$  & $1 - \epsilon$  \\ \hline
\end{tabular}
\end{center}
\end{table}

\subsection{Objective}
\label{sec:objective}
The first part of our objective is to maximize the likelihood of the observed constraints given the
instances, i.e.,
\begin{equation}
\renewcommand{\arraystretch}{1.2}
\label{eq:mle}
%\hspace{-0.05cm}
  \begin{array}{r  l}
 \displaystyle \max_{W} ~  \Phi(L| X_I;W) = & \hspace{-0.1cm} \frac{1}{M} \displaystyle \log  P(L|X_I;W) \\
                        ~                 = & \hspace{-0.1cm} \frac{1}{M} \displaystyle \log \sum_{Y_I} P(L,Y_I|X_I;W) ~,
  \end{array}
\end{equation}
where $I$ indexes the constrained instances  as defined in Section \ref{sec:notation}, and $\frac{1}{M}$ is a normalization constant.

To reduce overfitting, we add the standard L-2 regularization for the logistic model, namely,
\begin{equation}
\nonumber
\label{eq:L2}
   R(W) =  \sum_{k} {\tilde{w}_k}^T \tilde{w}_k  ~,
\end{equation}
where each $\tilde{w}_k$ is a vector obtained by replacing the bias term in $w_k$ with $0$.

In addition to satisfying the constraints, we also expect the clustering solution to
separate the clusters with large margins. This objective can be captured by
minimizing the conditional entropy of instance cluster labels given the observed  features \cite{yves05NIPS}.
Since the cluster information about constrained instances is captured by Eq.\ (\ref{eq:mle}),
we only impose such entropy minimization on the unconstrained instances, i.e.,
\begin{equation}
\nonumber
\label{eq:separation_term}
%\begin{array}{rl}
    H(Y_U|X_U;W) = \frac{1}{|U|}\sum_{i\in U}H\left[ P(y_i|x_i;W) \right] ~.
    %      = &\displaystyle - \frac{1}{|U|}\sum_{i\in U} \sum_{k=1}^K p_{ik}\log p_{ik}
%\end{array}
\end{equation}
%where $p_{ik} \equiv P(y_i = k | x_i; W)$.

Adding the above terms together, our objective is
\begin{equation}
\label{eq:obj1}
 \max_{W} ~~  \Phi(L| X_I;W)  - \tau H(Y_U|X_U;W) -  \lambda R(W) ~.
\end{equation}

In some cases, we may also wish to maintain a balanced distribution across different clusters.
This can be achieved by maximizing the entropy of the estimated marginal distribution of
cluster labels \cite{gomes10NIPS}, i.e.,
\begin{equation}
\nonumber
\label{eq:balance_term}
\begin{array}{l}
 \hspace{-0.1cm} H(\hat{y}|X;W)=  -\sum_{k=1}^K \hat{p}_k \log \hat{p}_k ~,\\
  % \hspace{0.8cm} \text{with} \quad \hat{p}_k \equiv \hat{P}(y=k|X;W) = \frac{1}{N} \sum_{i=1}^N p_{ik}.
\end{array}
\end{equation}
where we denote the estimated marginal probability as $\hat{p}_k = \hat{P}(y=k|X;W) = \frac{1}{N} \sum_{i=1}^N p_{ik}$
and $p_{ik} = P(y_i = k | x_i; W)$.

In cases where  balanced clusters are desired,  our objective is formulated as
\begin{equation}
\label{eq:obj}
\hspace{-0.1cm}
 \begin{array}{l}
   \displaystyle  \max_{W} ~  \Phi(L| X_I;W) -  \lambda R(W)  \\
   \displaystyle \hspace{1.5cm} +  ~ \tau \left[H(\hat{y}|X;W) - H(Y_U|X_U;W)\right] ~ ,
  \end{array}
\end{equation}
where we use the same coefficient $\tau$ to control the  enforcement of the cluster separation
and cluster balance terms, since they are roughly at the same scale.

The two objectives (\ref{eq:obj1}) and (\ref{eq:obj}) are non-concave, and optimization  generally
can only be guaranteed to reach a local optimum. In the next section, we present a variational EM
solution and discuss an effective initialization strategy.

\section{Optimization}
\label{sec:optimization}
Here we consider optimizing the objective in Eq.\ (\ref{eq:obj}), which enforces cluster balance.
The objective (\ref{eq:obj1}) is simpler and can be optimized following the same procedure
by simply removing the corresponding terms employed for cluster balance.

Computing the log-likelihood Eq.\ (\ref{eq:mle}) requires marginalizing
over hidden variables $Y_I$.  Exact inference may be feasible when
the constraints are highly separated or the number of constraints is
small, as this may produce a graphical model with low
tree-width. As more $y$'s are related to each other via
constraints, marginalization becomes more expensive to compute,
and it is in general intractable. For this reason,
we use the variational EM algorithm for optimization.

Applying Jensen's inequality, we obtain the lower bound of the objective as follows
%\begin{equation}
%\label{eq:lb}
%\renewcommand{\arraystretch}{1.3}
%\begin{array}{rl}
%\hspace{-0.2cm} LB = &\hspace{-0.2cm} \frac{1}{M}  \sum_{Y_I}Q(Y_I) \log \left[ \frac{P(Y_I,L|X_I;W)}{Q(Y_I)} \right] -  \lambda R(W)  \\
%\hspace{-0.2cm}      &\hspace{-0.2cm}              + ~ \tau [H(\hat{y}|X;W) - H(Y_U|X_U;W)]  ~,
%\end{array}
%\end{equation}
\begin{equation}
\label{eq:lb}
\renewcommand{\arraystretch}{1.3}
\begin{array}{rl}
\hspace{-0.2cm} LB = &\hspace{-0.2cm} \frac{1}{M}  E_{Q(Y_I)} \left[ \log ( \frac{P(Y_I,L|X_I;W)}{Q(Y_I)}) \right] -  \lambda R(W)  \\
\hspace{-0.2cm}      &\hspace{-0.2cm}              + ~ \tau [H(\hat{y}|X;W) - H(Y_U|X_U;W)]  ~,
\end{array}
\end{equation}
where $Q(Y_I)$ is a variational distribution. In variational EM, such lower bound is
maximized alternately in the E-step and M-step respectively\cite{bishopPRML}. In each E-step,
%for a fixed parameter $W$, the {\it LB} is maximized when the
we aim to find a tractable distribution $Q(Y_I)$ such that the Kullback-Leibler divergence
between $Q(Y_I)$ and the posterior distribution $P(Y_I|L,X_I;W)$ is minimized.
%, and we aim to find such a distribution $Q(Y_I)$
Given the current $Q(Y_I)$, each M-step finds the new $W$ that maximizes the {\it LB}.
Note that in the objective (and the {\it LB}), only the likelihood term is relevant to the E-step. The
other terms are only used in solving for $W$ in the M-steps.

\subsection{Variational E-Step}
\label{sec:estep}
\begin{table}
\caption{\label{tb:weight_fun}The values of $\tilde{Q}(l_t|y_i = k)$, $i \in I_t$.
   For simplicity, we denote $q_{jk}\equiv q(y_{t_j} = k)$ and $q_{j\bar{k}}\equiv q(y_{t_j}\neq k)$. }
\centering
\renewcommand{\arraystretch}{1.5}
\begin{tabular} {|l|p{1.5cm}p{1.5cm}p{3cm}|}
\hline
 {Cases}   & $l_t =  yes$                    & $ l_t =  no $              & ~~~~$l_t =  dnk $\\   \hline  \hline
$i=t_{1}$ & $q_{2k}q_{3 \bar{k}}$         & $q_{2 \bar{k}}q_{3 k}$        & $\displaystyle 1-q_{2k}q_{3 \bar{k}} - q_{2 \bar{k}}q_{3 k}$  \\
%    &   & \\ % \hline
$i=t_{2}$ & $q_{1k}q_{3 \bar{k}}$         & ${\sum \limits_{u\neq k}} q_{1u}q_{3u}$ & $1-q_{1k}q_{3 \bar{k}} -\hspace{-0.1cm} {\sum \limits_{u\neq k}} q_{1u}q_{3u}$ \\% \hline
%    &   & \\
$i=t_{3}$ & ${\sum \limits_{u\neq k}} q_{1u}q_{2u}$ & $q_{1k}q_{2 \bar{k}}$         & $1-q_{1k}q_{2 \bar{k}} -\hspace{-0.1cm} {\sum \limits_{u\neq k}} q_{1u}q_{2u}$ \\
\hline
\end{tabular}
\end{table}
We use  mean field inference \cite{Saul96, fox2012tutorial} to approximate the posterior distribution
in part due to its ease of implementation and convergence properties \cite{benaim2011mean}.
Mean field restricts the variational distribution $Q(Y_I)$ to the tractable fully-factorized family
$\label{eq:approxi}  Q(Y_I) = \prod_{i \in I} q(y_i)$,
and finds the $Q(Y_I)$ that minimizes the KL-divergence $KL[Q(Y_I)||P(Y_I|L,X_I;W)]$.
The optimal $Q(Y_I)$ is obtained by iteratively updating each $q(y_i)$ until $Q(Y_I)$ converges.
The update equation is
%Romer removed this {fox2012tutorial} for an earlier reference
%Yuanli: The earlier reference provides an example of using mean-field in sigmoid
%        belief network, but it didn't explicitly provide the update equation as follows.
%        So I think it is better to keep both.
%\begin{equation}
%\label{eq:update_logq}
%    \log q(y_i) \varpropto E_{Q(Y_{I\backslash i})}[\log P(X_I,Y_I,L)], \quad i \in I \\
%\end{equation}
%where $Q(Y_{I\backslash i}) = \prod_{j\in I, j\neq i} q(y_j)$ .
%Correspondingly,
\begin{equation}
\label{eq:update_q}
    q(y_i) = \frac{1}{Z} \exp\{E_{Q(Y_{I\backslash i})}[\log P(X_I,Y_I,L)] \} ~,
\end{equation}
where $Q(Y_{I\backslash i}) = \prod_{j\in I, j\neq i} q(y_j)$,
and $Z$ is a normalization factor to ensure $\sum_{y_i} q(y_i)=1$.
In the following, we derive a closed-form update for this optimization problem.

Applying the model independence assumptions, the expectation
%$E_{Q(Y_{I\backslash i})}\left[\log P(X_I,Y_I,L)\right]$
term in Eq.\ (\ref{eq:update_q}) is simplified to
%\begin{equation}
%\label{eq:estep_update}
%   \sum_{t: i \in I_t} E_{Q(Y_{I_t \backslash i})} [\log P(l_t|y_i)] + \log P(y_i|x_i;W) + const,
%\end{equation}
%%\begin{equation}
%%\label{eq:estep_update}
%%%\hspace{-0.21cm}
%%\renewcommand{\arraystretch}{1.4}
%%\begin{array}{l}
%%%~~~   \displaystyle   E_{Q(Y_{I\backslash i})}\left[\log P(X_I,Y_I,L)\right] \\
%%  \displaystyle \hspace{-0.2cm} \sum_{Y_{I\backslash i}} Q(Y_{I\backslash i})
%%                    \left(\sum_{j\in I} \log  P(y_j|x_j;W) + \sum_{t = 1}^M  \log P(l_t| Y_t,y_i) \right) \\
%%=  \displaystyle  \hspace{-0.12cm} \sum_{t: i \in I_t} E_{Q(Y_{I_t \backslash i})} [\log P(l_t|y_i)] + \log P(y_i|x_i;W) + const,
%%               % \sum_{Y_{I_t \backslash i}} Q( Y_{I_t \backslash i}) \log P(l_t|Y_t,y_i) +  c,
%%\end{array}
%%\end{equation}
\begin{equation}
\label{eq:estep_update}
\renewcommand{\arraystretch}{1.4}
\hspace{-0.2cm}
\begin{array}{rl}
   & \hspace{-0.3cm} E_{Q(Y_{I\backslash i})}
                     [ {\sum \limits_{t = 1}^M}  \log P(l_t| Y_t) +{\sum \limits_{j\in I}} \log  P(y_j|x_j;W) + \log P(X_I)]\\
=  & \hspace{-0.3cm} {\sum \limits_{t: i \in I_t}} E_{Q(Y_{I_t\backslash i})} [\log P(l_t|Y_t)] + \log P(y_i|x_i;W) + const,
               % \sum_{Y_{I_t \backslash i}} Q( Y_{I_t \backslash i}) \log P(l_t|Y_t,y_i) +  c,
\end{array}
\end{equation}
where $I_t \backslash i$ is the set of indices in $I_t$ except for $i$, and $const$ absorbs all the terms that
are constant with respect to $y_i$. The first term in  (\ref{eq:estep_update})  sums over the expected log-likelihood of observing
each $l_t$ given the fixed $y_i$. To compute the expectation, we first let $\tilde{Q}(l_t|y_i)$  be the probability that the observed
$l_t$ is consistent with  the $Y_t$  given a fixed $y_i$. That is, $\tilde{Q}(l_t|y_i)$ is the probability
for all possible assignments of $Y_t$ given a fixed $y_i$, such that $P(l_t|Y_t) = 1-\epsilon$
according to Table \ref{tab:label_dist}. The $\tilde{Q}(l_t|y_i)$ can be computed
straightforwardly as in Table \ref{tb:weight_fun}. Then each of the
expectations in (\ref{eq:estep_update}) is computed as
\[
E[\log P(l_t|y_i)] =[1-\tilde{Q}(l_t|y_i)] \log \frac{\epsilon}{2} + \tilde{Q}(l_t|y_i)\log (1-\epsilon).
\]
From the above, the update Eq.\ (\ref{eq:update_q}) is derived as
\begin{equation}
%\nonumber
\label{eq:estep_q}
 q(y_i) = \displaystyle  \frac{ \displaystyle \alpha^{F(y_i) } P(y_i|x_i;W) }
                  { \sum_{y_i} \displaystyle \alpha^{F(y_i) } P(y_i|x_i;W)}, ~~ \text{with} ~~ \alpha = \frac{2(1-\epsilon)}{\epsilon}  ~,
\end{equation}
where $F(y_i) = \sum_{t: i \in I_t} \tilde{Q}(l_t|y_i)$.

The term $F(y_i)$ can be interpreted  as measuring the compatibility of each assignment of $y_i$ with respect
to  the constraints and the other $y$'s. In Eq.\ (\ref{eq:estep_q}), $\alpha$ is controlled
by the parameter $\epsilon$. When $\epsilon \in (0, \frac{2}{3})$,  $\alpha > 1$ and
the update allows more compatible assignments of $y_i$, i.e., the ones with higher $F(y_i)$,
to have larger $q(y_i)$. When $\epsilon = \frac{2}{3}$, the constraint
labels are regarded as uniformly distributed regardless of the instance cluster labels, as can be seen
from Table \ref{tab:label_dist}. In this case, $\alpha = 1$ and each $q(y_i)$ is directly
set to the conditional probability $P(y_i|x_i;W)$.
This naturally reduces our method to learning without constraints.
Clearly, when $\epsilon$ is smaller, the constraints are harder and the updates will push $q(y_i)$ to more
extreme distributions. %to favor assignments that are consistent with the constraints.
Note that the values of $\epsilon \in (\frac{2}{3},1)$ cause $\alpha < 1$, which
will lead to results that contradict the constraints, and are generally not desired.

%\subsubsection{Special Case: Hard Constraints}

{\it Special Case: Hard Constraints.}
In the special case where $\epsilon=0$ and $\alpha =\infty$, $P(l_t|Y_t)$ essentially
reduces to the deterministic model described in Eq.\ (\ref{eq:triple}), allowing our model
to incorporate {\it hard} constraints. The update equation of this case
can also be addressed similarly to Eq.\ (\ref{eq:estep_q}). In this case,
$q(y_i)$ is non-zero only when the value of $F(y_i)$ is the maximum
among all possible assignments of $y_i$. Thus, the update equation
is reduced to a max model. More formally, we define the max-compatible label set
for each instance $x_i$ as
\begin{equation*}
    Y_i = \{1\leq k \leq K: F(y_i = k) \geq F(y_{i} = k'), \forall ~ k' \neq k \}.
\end{equation*}
Namely, each $Y_i$ contains the most compatible assignments for $y_i$ with respect to the constraints.
Then the update equation becomes
\begin{equation}
%\nonumber
\label{eq:estep_q_hard}
 q(y_i) = \displaystyle  \left\{
            \begin{array} {c c }
                 \displaystyle \frac{ P(y_i|x_i;W) } { \sum_{y'_i \in Y_i} P(y'_i|x_i;W)}, & \text{if} ~~ y_i \in Y_i~, \\
                                0, & ~~\text{o.w.}
             \end{array}
             \right.
\end{equation}

\subsection{M-Step}
\label{sec:mstep}
The M-step searches for the parameter $W$ that maximizes the $LB$.
Applying the independence assumptions again and ignoring all the terms
that are constant with respect to $W$, we obtain the following objective
\begin{equation}
\nonumber
\label{eq:maxlb}
\begin{array}{rl}
 \displaystyle \max_{W} ~ J = &\hspace{-0.2cm} \frac{1}{M} \displaystyle  \sum_{Y_I} Q(Y_I) \log P(Y_I|X_I;W) -  \lambda R(W) \\
                              &\hspace{-0.2cm}  + \tau \left[H(\hat{y}|X;W) - H(Y_U|X_U;W)\right] . \\
\end{array}
\end{equation}

This objective is non-concave and a local optimum can be found via gradient
ascent. We used L-BFGS \cite{BFGS} in our experiments.
%\footnote{We use the L-BFGS algorithm.}
%The implementation is downloaded from \url{http://www.di.ens.fr/~mschmidt/Software/minFunc.html}}.
The derivative of $J$  w.r.t.\ $W$ is
\begin{equation}
\label{eq:mstep_w}
 \nonumber
\renewcommand{\arraystretch}{1.8}
\begin{array}{rl}
 \frac{\partial J}{\partial W} =
   & \hspace{-0.2cm} \frac{1}{M}         \sum_{i \in I} (Q_i - P_i)  x_i^T  - 2 \lambda \tilde{W} \\
   & \hspace{-0.2cm} + \frac{\tau}{|U|}  \sum_{j \in U}  \sum_{k} (\mathbf{1}_k - P_j) p_{jk} \log p_{jk}  x_j^{T} \\
   & \hspace{-0.2cm} - \frac{\tau}{N}    \sum_{n=1}^N  \sum_{k} (\mathbf{1}_k - P_n) p_{nk} \log \hat{p}_{k} x_n^{T}  ~~,
\end{array}
\end{equation}
where $P_i = [p_{i1}, \ldots, p_{iK}]^T$,
%$\hat{p}_k$ is defined at Eq.\ (\ref{eq:balance_term}),
$Q_i = [q_{i1}, \ldots, q_{iK}]^T$ with $q_{ik} = q(y_i = k)$,
$\tilde{W} = [\tilde{w}_1, \ldots, \tilde{w}_K]^T$, and $\mathbf{1}_k$ is
a $K$-dimensional vector that contains the value $1$ on
the $k$-{th} dimension and $0$ elsewhere.

The above derivations use a linear model for $P(y|x;W)$,
and thus the learned DCRC is also linear. However, all of the results can be easily generalized
to using kernel functions, allowing DCRC to find non-linear separation boundaries.

\subsection{Complexity and Initialization}
\label{sec:initialization}
In each E-step, the complexity is $\mathcal{O}(\gamma K |I|)$,
where $\gamma$ is the number of mean-field iterations for $Q(Y_I)$ to converge.
%We empirically observed that $Q(Y_I)$ converges very fast, especially in later EM iterations.
In the M-step, the complexity of computing the gradient of $W$ in each
L-BFGS iteration is $\mathcal{O}(NKD)$.

%In practice, for the mean field approximation in the E-step, we can
%use a fixed number of update iterations as an additional termination
%condition. We empirically observed that mean field converges very fast, especially
%in later EM iterations. Similarly, in the M-step the L-BFGS optimization usually
%converges with fewer iterations in the later EM iterations. Thus, a
%fixed number of update completion was also employed as an appropriate termination
%condition.

Although mean-field approximation is guaranteed to converge, in the first few E-steps
it is not critical to achieve a very close approximation.
In practice, we can run mean-field update up to a fixed number of
iterations (e.g., 100). We empirically observe that the approximation
still converges very fast in later EM iterations.
%,even if the approximation in the first few iterations is not optimal.
Similarly, we observe in the M-step that the L-BFGS optimization usually converges with very few iterations
in the later EM runs, and a completion of a fixed number of iterations for L-BFGS is
also sufficient in the first few M-steps.

The EM algorithm is generally sensitive to the initial parameter values.
Here we first apply Kmeans and train a supervised logistic classifier with
the clustering results. The learned weights are then used as the starting point of DCRC.
Empirically we observe that such initialization typically allows
DCRC to converge  %(the change of $W$ below a threshold)
within $100$ iterations.

\section{Experiments}
\label{sec:experiments}
\begin{figure*}[ht!]
\centering
     \subfigure[Ionosphere]{ \includegraphics[width=0.46\textwidth]{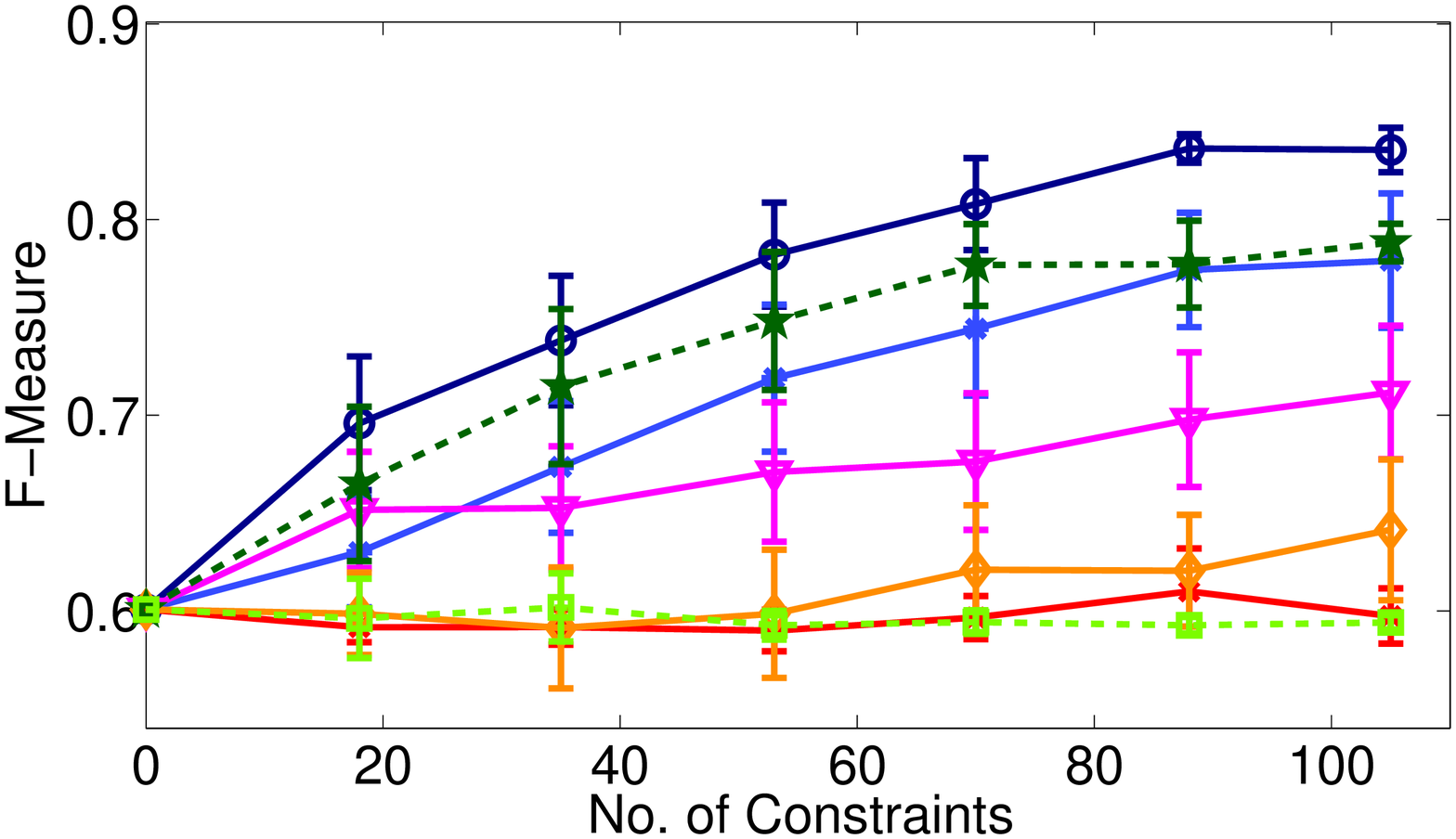} }
     \hspace{-0.5cm}
     \subfigure[Pima]{ \includegraphics[width=0.46\textwidth]{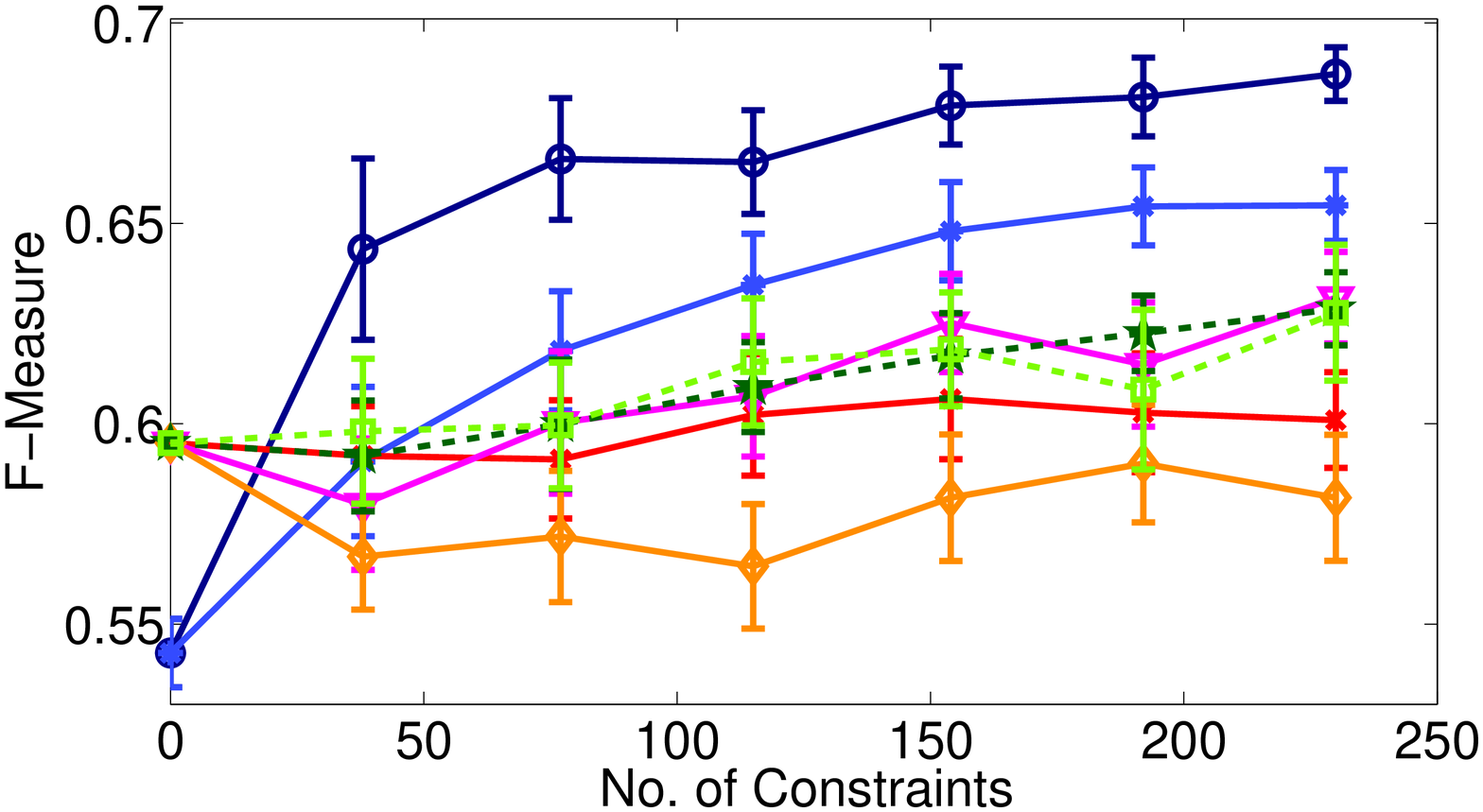} }
     \subfigure[Balance-scale]{ \includegraphics[width=0.46\textwidth]{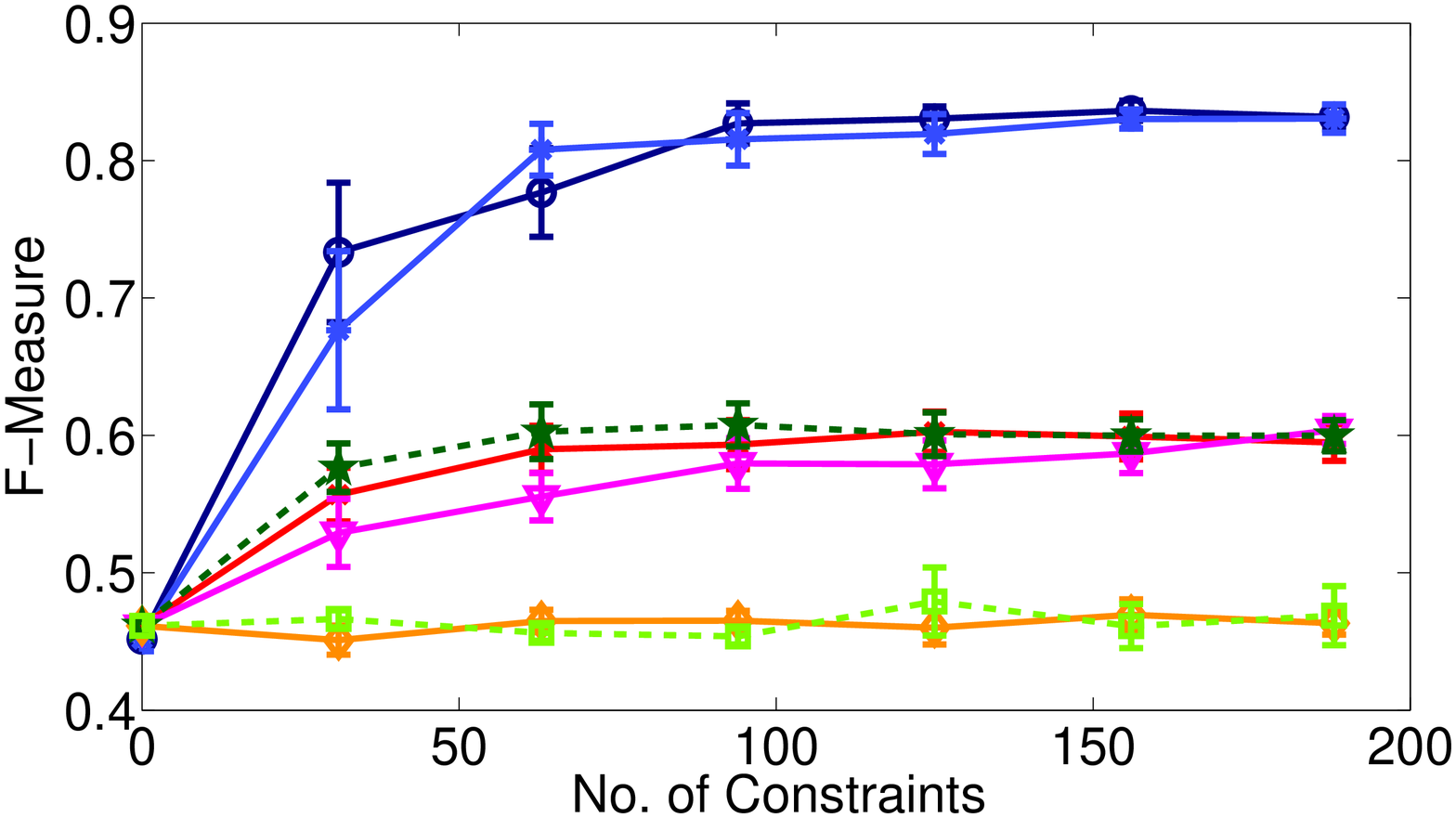} }
     \hspace{-0.5cm}
     \subfigure[Digits-389]{ \includegraphics[width=0.46\textwidth]{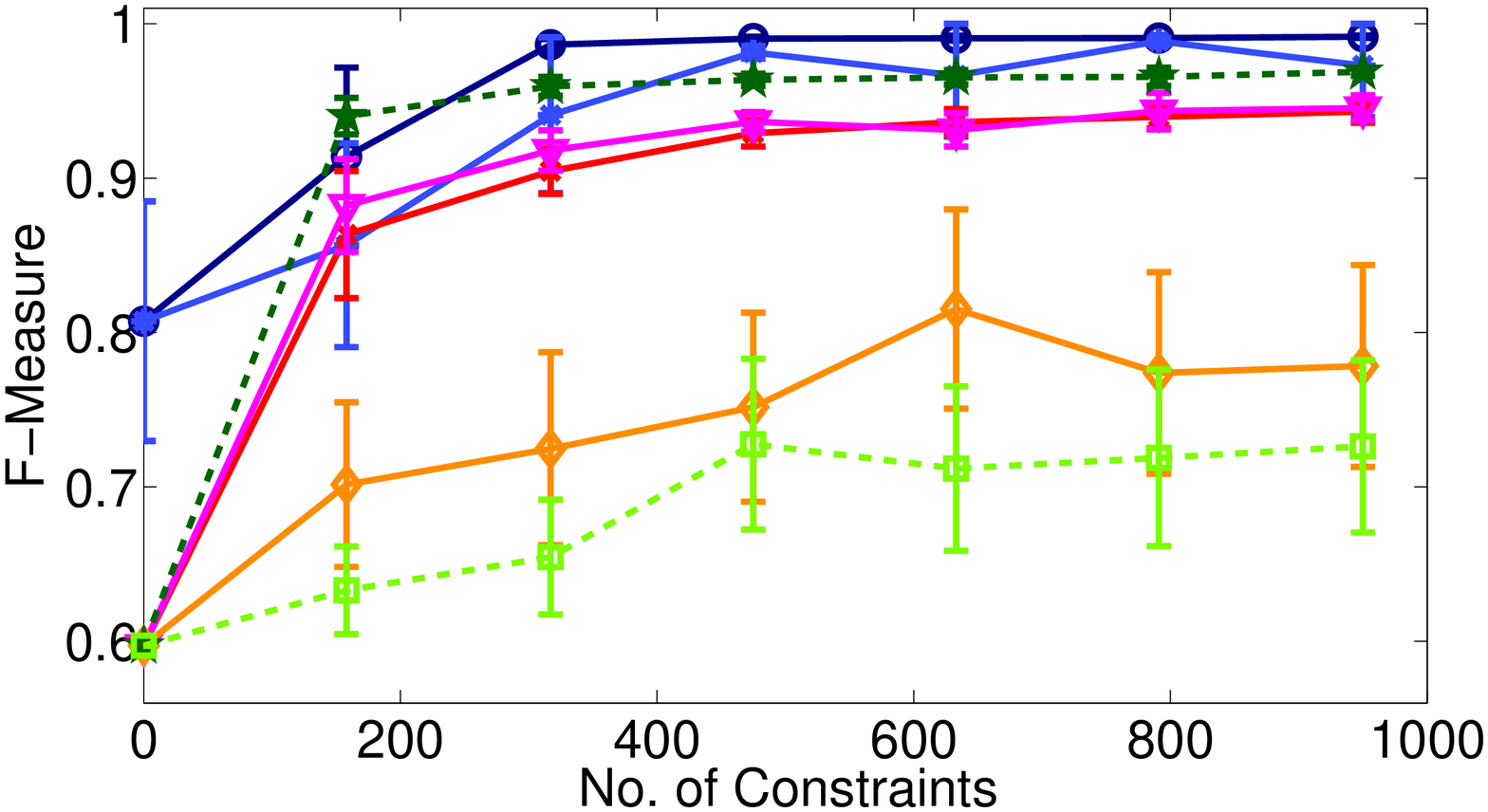} }
   %  \subfigure[Letters-IJ]{ \includegraphics[width=0.46\textwidth]{figure/letters-IJ_F1_cfi.eps} }
%     \hspace{-0.5cm}
     \subfigure[Letters-IJLT]{ \includegraphics[width=0.46\textwidth]{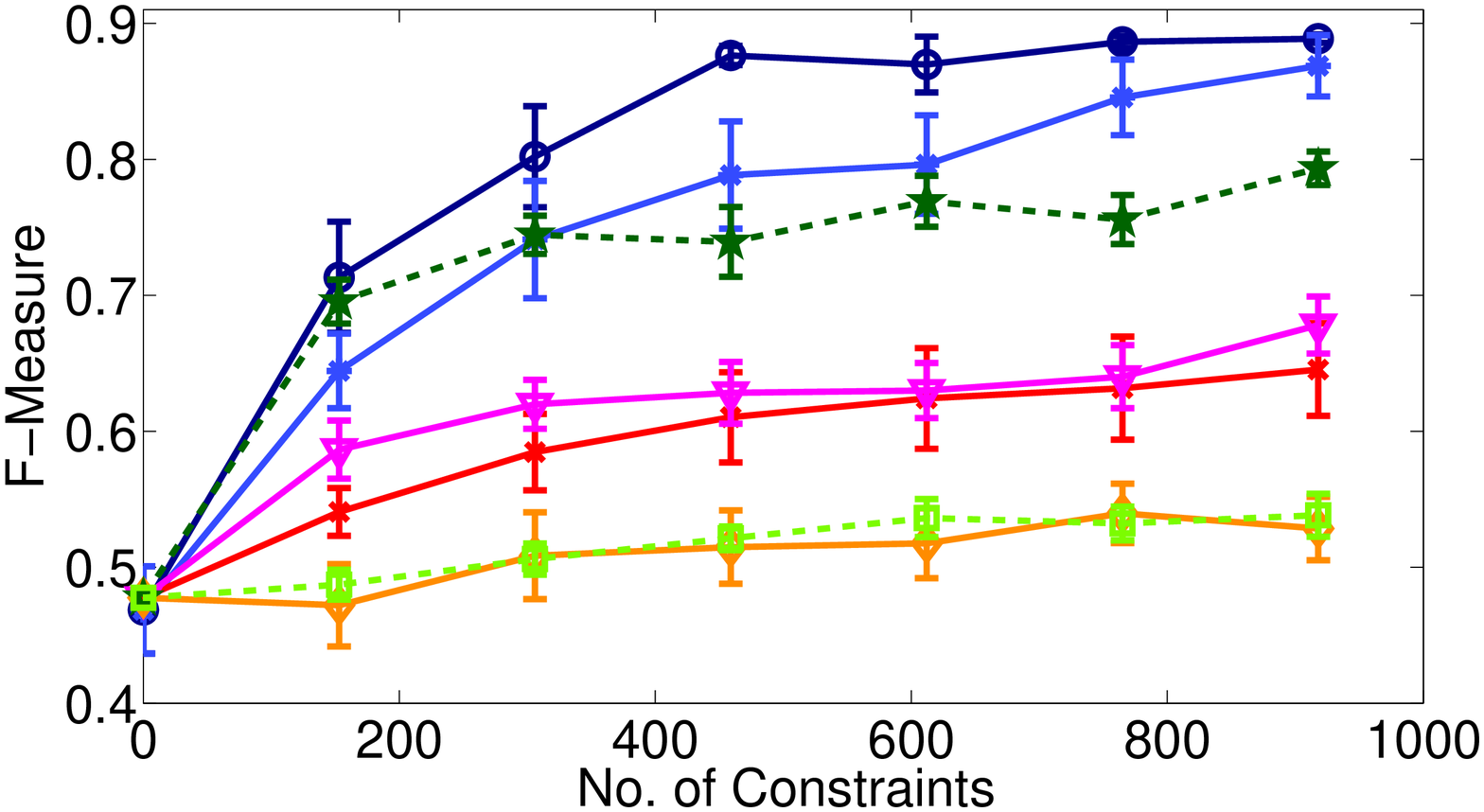} }
    % \subfigure[Birdsong-3Class]{ \includegraphics[width=0.46\textwidth]{figure/bird3Class_F1_cfi.eps} }
     \hspace{-0.5cm}
     \subfigure[MSRCv2]{ \includegraphics[width=0.46\textwidth]{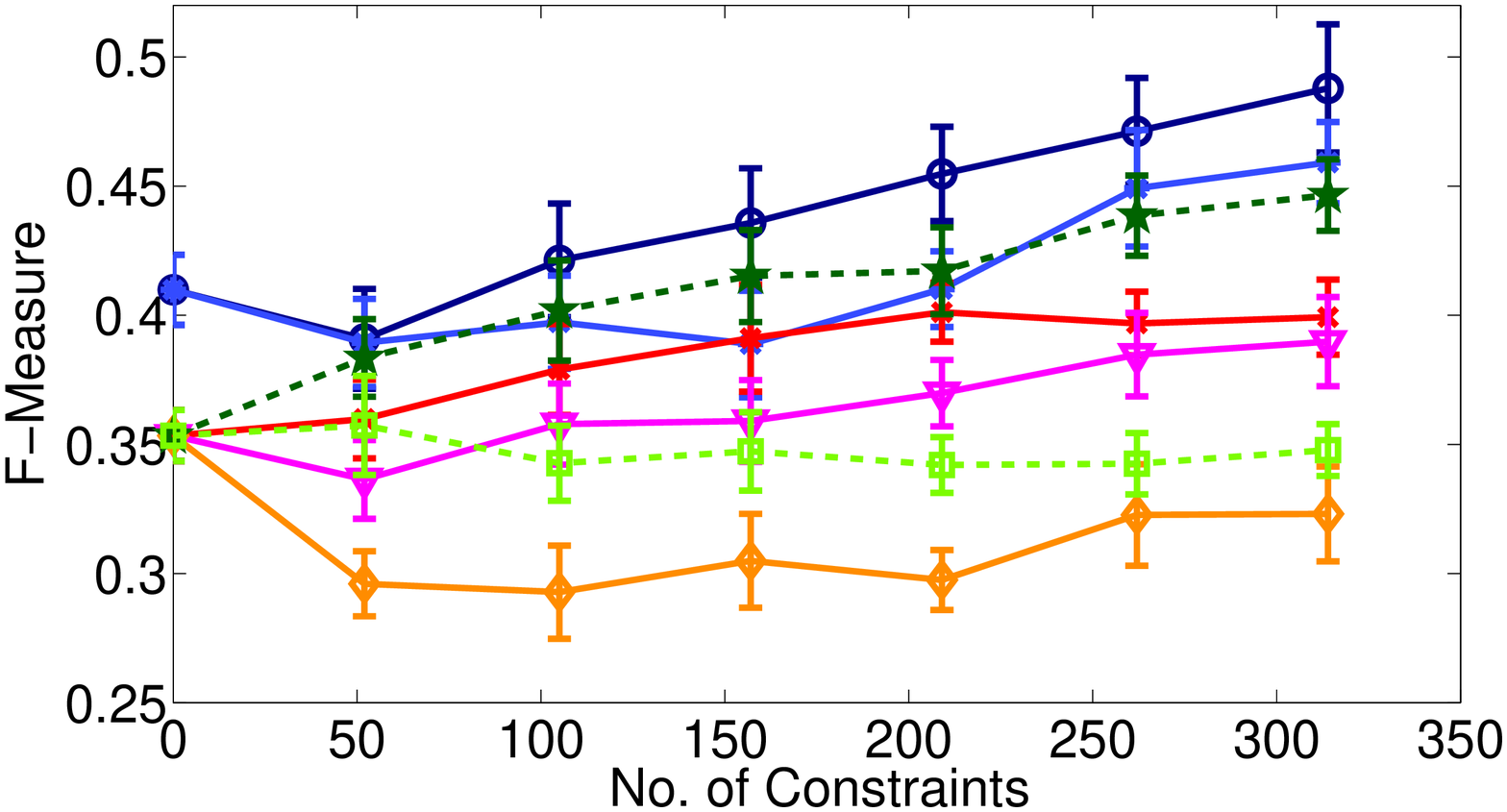} }
     \subfigure[Stonefly9]{ \includegraphics[width=0.46\textwidth]{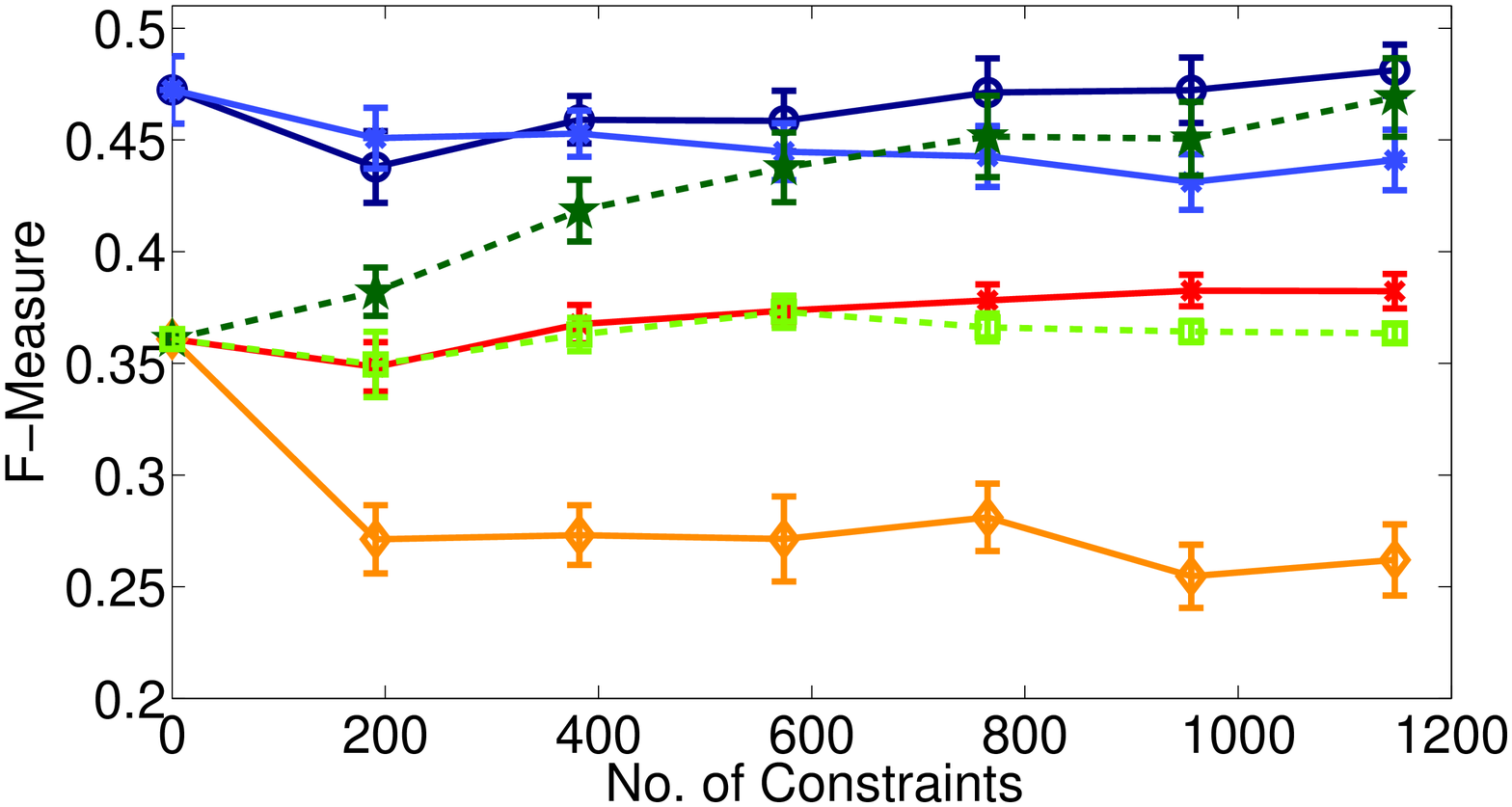} }
     \hspace{-0.5cm}
     \subfigure[Birdsong]{ \includegraphics[width=0.46\textwidth]{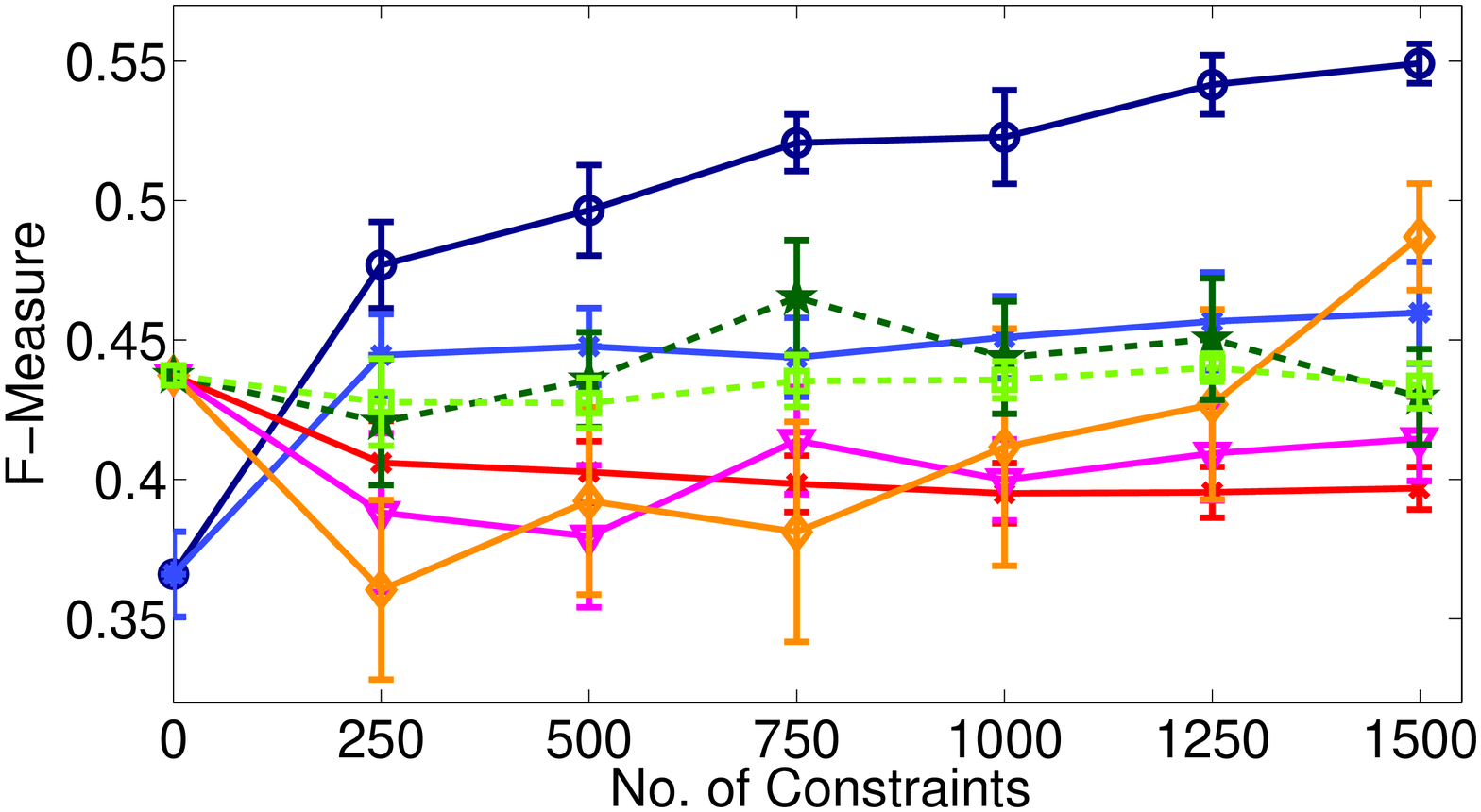} }
     \subfigure{ \includegraphics[width=0.9\textwidth]{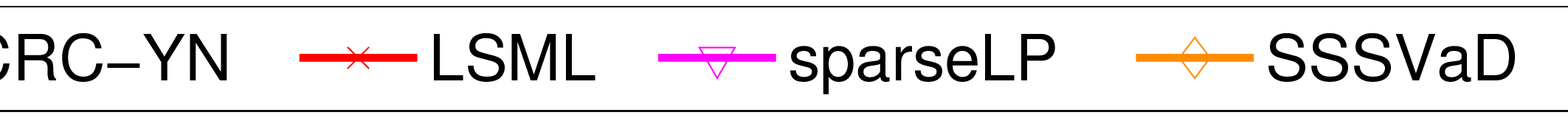} }
\caption{\label{fig:constr}(Best viewed in color.) The F-measure as a function of
        number of relative constraints.  Results are averaged over 20 runs with independently
        sampled constraints. Error bars are shown as 95\% confidence intervals. }
\end{figure*}

%\begin{figure*}[ht!]
%\centering
%\begin{tabular}{cc}
%     \includegraphics[width=0.49\textwidth]{figure/ionosphere_F1_cfi.eps} &
%     \hspace{-0.5cm}
%     \includegraphics[width=0.49\textwidth]{figure/pima_F1_cfi.eps} \\
%      (a) Ionosphere & (b) Pima  \\
%     \includegraphics[width=0.49\textwidth]{figure/balance-scale_F1_cfi.eps}  &
%     \hspace{-0.5cm}
%     \includegraphics[width=0.49\textwidth]{figure/digits389_F1_cfi.eps} \\
%      (c) Balance-scale  & (d) Digits-389 \\
%     \includegraphics[width=0.49\textwidth]{figure/letters-IJLT_F1_cfi.eps} &
%     \hspace{-0.5cm}
%     \includegraphics[width=0.49\textwidth]{figure/msrcv2Sub_F1_cfi.eps}  \\
%      (e) Letters-IJLT & (f) MSRCv2 \\
%     \includegraphics[width=0.495\textwidth]{figure/stonefly9_F1_cfi.eps}  &
%     \hspace{-0.5cm}
%     \includegraphics[width=0.49\textwidth]{figure/bird_F1_cfi.eps}  \\
%     (g) Stonefly9 &  (h) Birdsong \\ &\\
%     \multicolumn{2}{c}{~\includegraphics[width=0.96\textwidth]{figure/legend.eps} }
%\end{tabular}
%\caption{\label{fig:constr}(Best viewed in color.) The F-measure as a function of  number of constraints.
%    %For each dataset,the number of constraints ranges from 0 to 30\% times of the number of instances in the dataset.
%    Results are averaged over 20 runs with independently sampled constraints.
%    Error bars are 95\% confidence interval. }
%\end{figure*}
%

In this section, we experimentally examine the effectiveness of our model in utilizing relative constraints
to improve clustering. We first evaluate all methods on both UCI and other real-world
datasets with noise-free constraints generated from true class labels.
We then present a preliminary user study where we ask users to label constraints
and evaluate all the methods on these human-labeled (noisy) constraints.

\subsection{Baseline Methods and Evaluation Metric}
We compare our algorithm with existing methods that consider relative constraints or pairwise constraints.
The methods employing pairwise constraints are {\it Xing}'s
method \cite{xing2003NIPS} (distance metric learning for a diagonal matrix)
%\footnote{We use metric learning to learn a diagonal matrix.}
%, since we observed that it is generally faster than learning a full matrix and also performs better on most of our datasets.},
%{\bf MPCKmeans} \cite{basu04},
and {\it ITML} \cite{davis07ICML}. These are the state-of-the-art methods that are
usually compared in the literature and have publicly available source code.

For methods considering relative constraints,
we compare with: 1) {\it LSML} \cite{liu2012ICDM},
%\footnote{The prior is set to the Euclidean distance.},
a very recent metric learning method studying relative constraints (we use Euclidean distance as the prior); 2) {\it SSSVaD} \cite{kumar08TKDE},
a method that directly finds clustering solutions with relative constraints;
and 3) {\it sparseLP} \cite{rosales2006KDD}, an earlier method that hasn't been extensively compared.
We also experimented with a SVM-style method proposed in \cite{schultz2003NIPS}
and observed that its performance is generally worse. % than these baselines.
Thus, we do not report the results on this method.

Xing's method, ITML,  LSML, and sparseLP are metric learning techniques.
Here we apply Kmeans  with the learned metric (50 times) to form cluster assignments, and
the clustering solution with the minimum mean-squared error is chosen.
%\textcolor[rgb]{1.00,0.00,0.00}{(with multiple random initialization?)}
%%Yuanli: No. For each run of Kmeans, we used random initialization once. Overall Kmeans is run 50 times.
%%        I think it is fine to just say that we used Kmeans.

We evaluated the clustering results based on the ground-truth class labels using
\emph{pairwise F-measure} \cite{basu04},  \emph{Adjusted Rand Index}
and \emph{Normalized Mutual Information}. The results are highly similar with different measures,
thus we only present the F-Measure results.

%The F-measure evaluates clustering by considering same-cluster instance pairs:
%\begin{eqnarray*}
%    &&\hspace{-0.4cm}  Precision = \displaystyle  \frac{\# PairsCorrectlyPredictedInSameCluster}{\# TotalPairsPredictedInSameCluster} \\
%    &&\hspace{-0.4cm}  Recall =    \displaystyle  \frac{\# PairsCorrectlyPredictedInSameCluster}{\# TotalPairsInSameCluster} \\
%    &&\hspace{-0.4cm}  F\text{-}Measure = \displaystyle  \frac{2\times Precision \times Recall} {Precision  + Recall}
%\end{eqnarray*}

\subsection{Controlled Experiments}
\label{sec:noisefree}
In this set of experiments, we use simulated noise-free constraints to evaluate all the methods.
\subsubsection{Datasets}
\begin{table}
\caption{\label{tab:dataset}Summary of Dataset Information}
\centering
\begin{tabular}{|c|c|c|c|}
\hline
    Dataset         &  \#Inst.  &  \#Dim.   & \#Cluster \\ \hline \hline
    Ionosphere      &  351      &   34      &    2  \\% \hline
    Pima            &  768      &   8       &    2  \\% \hline
    Balance-scale   &  625      &   4       &    3  \\% \hline
    Digits-389      &  3165     &  16       &    3  \\% \hline
%    Letters-IJ      &  1502     &  16       &    2  \\% \hline
%    Letters-IJLT    &  3059     &  16       &    4  \\% \hline
   Letters-IJLT     &  3059       &  16       &    4  \\% \hline
    MSRCv2          &  1046     &  48       &    6  \\% \hline
    Stonefly9        &  3824    &  285      &    9  \\% \hline
%    Birdsong-3Classes&  1913    &  38       &     3 \\  % \hline
%    Birdsong      &  4998       &  38       &    13 \\  \hline
   Birdsong    &  4998    &  38     &    13 \\  \hline
\end{tabular}
\end{table}
We evaluate all methods on five UCI datasets:
\emph{Ionosphere}, \emph{Pima}, \emph{Balance-Scale},  \emph{Digits-389}, and \emph{Letters-IJLT}.
We also use three extra real-world datasets: 1) a subset of image segments of the \emph{MSRCv2}
data\footnote{\url{http://research.microsoft.com/en-us/projects/ObjectClassRecognition/}},
%\footnote{The images are available at \url{http://research.microsoft.com/en-us/projects/ObjectClassRecognition/}.},
%We use the same features  extracted in \cite{briggs12KDD}.},
which contains the six largest classes of the image segments;
2) the HJA \emph{Birdsong} data \cite{briggs12KDD}, which contains automatically
extracted segments from spectrograms of birdsong recordings, and the goal is to identify the species for each segment;
%Here, each instance is a segment of the spectrogram for a 10-second birdsong recording, and we
%and the segments are labeled with the bird species for the corresponding utterance.
and 3) the \emph{Stonefly9} data\cite{stonefly09},
%\footnote{The dataset is downloaded
%from \url{http://web.engr.oregonstate.edu/~tgd/bugid/stonefly9/}. We used the SIFT descriptors of Harris and Hessian
%keypoints, Kadir-Brady salient regions, and shape features to create a dictionary of $540$ keywords using Kmeans,
%to represent the images using bag-of-words representation. The dimension is then reduced to $285$ via
%PCA to retain $75\%$ of the total variance.},
which contains insect images and the task is to identify the species of the insect for each image.
Table \ref{tab:dataset} summarizes the dataset information.
In our experiments, all features are standardized to have zero mean and unit standard deviation.

\subsubsection{Experimental Setup}
For each dataset, we vary the number of constraints from $0.05N$ to
$0.3N$ with a $0.05N$ increment, where $N$ is the total number of instances.
For each size, triplets are randomly generated and constraint labels are assigned
%based on instance class (cluster) labels
according to Eq.\ (\ref{eq:triple}).
% Such constraints are used to form the input of all the methods.
%We set $\epsilon = 0.05$ for our method ({\it i.e.,} allowing soft constraints) and evaluated it on two settings,
We evaluated our method in two settings, one with all constraints as input (shown as \emph{DCRC}), and the other
with only \emph{yes}/\emph{no} constraints  (shown as \emph{DCRC-YN}).
The baseline methods for relative constraints are designed for \emph{yes}/\emph{no} constraints only
and cannot be easily extended to incorporate \emph{dnk} constraints, so we drop the \emph{dnk}
constraints for these methods. To form the corresponding pairwise
constraints, we infer one ML and one CL constraints from each relative constraint with \emph{yes}/\emph{no} labels
(note that no pairwise constraints could be directly inferred from \emph{dnk} relative constraints).
%That is, for each constraint $(x_{t_1}, x_{t_2}, x_{t_3})$ with  \emph{yes}/\emph{no} label,
%we infer a ML constraint $(x_{t_1}, x_{t_2})$/$(x_{t_1}, x_{t_3})$ and a
%CL constraint $(x_{t_1}, x_{t_3})$/$(x_{t_1}, x_{t_2})$.
Thus, all the baselines use the same information as DCRC-YN, since no \emph{dnk}
constraints are employed by them.

%For all methods, we use five-fold cross-validation with the available constraints to select
%parameters that maximize the constraint prediction accuracy on the validation set.
%For both training and validation, all the methods are using the same set of folds.

We use five-fold cross-validation to tune parameters for all methods.
The same training and validation folds are used across all the methods (removing \emph{dnk}
constraints, or converting to pairwise constraints when necessary).
For each method, we select the parameters that maximize the averaged constraint prediction accuracy on the validation sets.
For our method, we search for the optimal $\tau \in \{0.5,1,1.5\}$ and $\lambda \in \{2^{-10},2^{-8},2^{-6},2^{-4},2^{-2}\}$.
We empirically observed that our method is very robust to the choice of $\epsilon$ when it is within the range $[0.05, 0.15]$.
Here we set $\epsilon = 0.05$ for this set experiments with the simulated noise-free constraints.
%For validation, instance labels are first assigned to the cluster with the highest $P(y|x;W)$.
%Constraint labels are then predicted based on the formed clusters.
%For all methods, the parameters with the best prediction accuracy on the validation constraints are chosen.
%The \emph{dnk} constraints are not used in the cross validation for baselines.
Experiments are repeated using 20 randomized runs, each with independently sampled constraints.
%For each run, the same constraint set is used to create inputs for all methods.

%The same sets of cross validation are used for baselines. Constraint labels are predicted based on distance comparisons in the
%validation of sparseLP and  LSML, and on output clustering solutions in that of SSSVaD.
%we first compute the distances $d(x_{i}, x_{j})$ and $d(x_{i}, x_{k})$,
%and then predict the label of constraint $(x_{i}, x_{j}, x_k)$ to \emph{yes} if $d(x_{i}, x_{j}) < d(x_{i}, x_{k})$,
%and to \emph{no} otherwise.
%Since \emph{dnk} constraints are not employed by these methods, they are not used in their cross validation.

\subsubsection{Overall Performance}
\label{sec:const_size}
Figure \ref{fig:constr} shows the performance of all methods with different number of constraints.
The sparseLP does not scale to the high-dimensional \emph{Stonefly9} dataset
and hence is not reported on this particular data.

From the results we see that DCRC consistently
outperforms all baselines on all datasets as the constraints increase,
demonstrating the effectiveness of our method.

Comparing DCRC with DCRC-YN, we observe that the additional \emph{dnk}
constraints provide substantial benefits, especially for datasets with
large number of clusters (e.g., \emph{MSRCv2}, \emph{Birdsong}). This is consistent
with our expectation because the portion of \emph{dnk} constraints
increases significantly when $K$ is large, leading to more information to be utilized by DCRC.

Comparing DCRC-YN with the baselines, we observe that DCRC-YN
achieves comparable or better performance even compared with the
best baseline ITML. This suggests that, with noise-free
constraints, our model is competitive with the state-of-the-art methods even without
considering the additional information provided by \emph{dnk} constraints.

\subsubsection{Soft Constraints vs.\ Hard Constraints}
%\begin{figure}[ht!]
%\centering
%     \subfigure[Ionosphere]{ \includegraphics[width=0.24\textwidth]{figure/ionosphere_F1_hard.eps} }
%     \hspace{-0.5cm}
%%     \subfigure[Balance-scale]{ \includegraphics[width=0.24\textwidth]{figure/balance-scale_F1_hard.eps} }
%%     \hspace{-0.5cm}
%%     \subfigure[Pima]{ \includegraphics[width=0.24\textwidth]{figure/pima_F1_hard.eps} }
%    % \hspace{-0.5cm}
%    % \subfigure[Digits-389]{ \includegraphics[width=0.24\textwidth]{figure/digits389_F1_hard.eps} }
%     %\subfigure[Letters-IJ]{ \includegraphics[width=0.24\textwidth]{figure/letters-IJ_F1_hard.eps} }
%     \subfigure[Letters-IJLT]{ \includegraphics[width=0.24\textwidth]{figure/letters-IJLT_F1_hard.eps} }
%    % \hspace{-0.5cm}
%%     \subfigure[Birdsong-3Class]{ \includegraphics[width=0.24\textwidth]{figure/bird3Class_F1_hard.eps} }
%     \hspace{-0.5cm}
%     \subfigure[Birdsong]{ \includegraphics[width=0.24\textwidth]{figure/bird_F1_hard.eps} }
%     %\subfigure[MSRCv2]{ \includegraphics[width=0.24\textwidth]{figure/msrcv2Sub_F1_hard.eps} }
%     \hspace{-0.5cm}
%     \subfigure[Stonefly9]{ \includegraphics[width=0.24\textwidth]{figure/stonefly9_F1_hard.eps} }
%\caption{\label{fig:hardConst} Comparison of DCRC using soft and hard constraints.
%   % Results are averaged over 20 runs with independently sampled constraints.
%   % Error bars are shown as mean and 95\% confidence interval.
%    }
%\end{figure}
\begin{figure}
\centering
\begin{tabular}{cc}
    \includegraphics[width=0.235\textwidth]{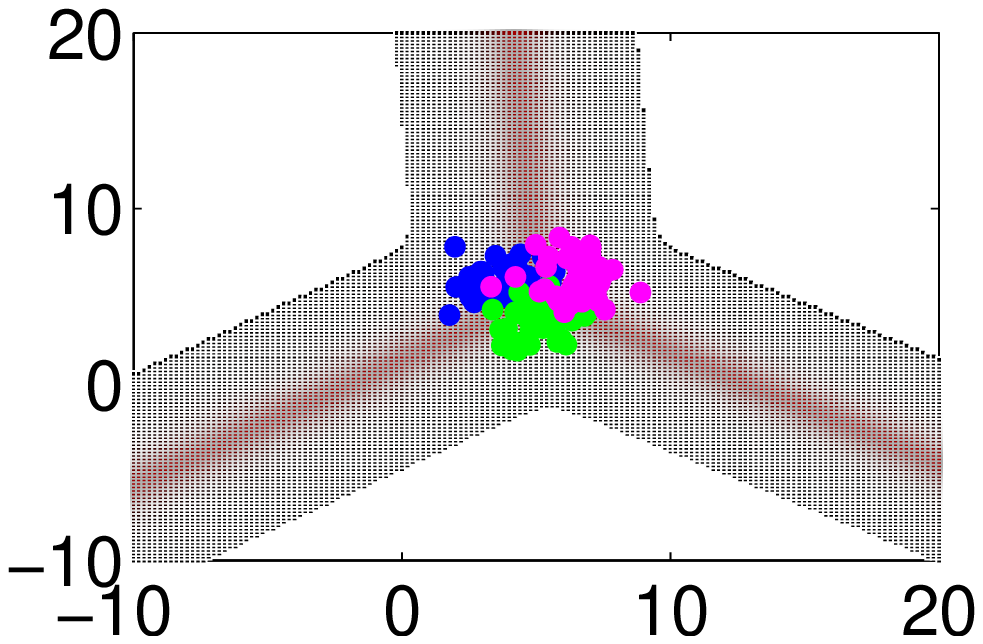}
    & \hspace{-0.1cm}
    \includegraphics[width=0.235\textwidth]{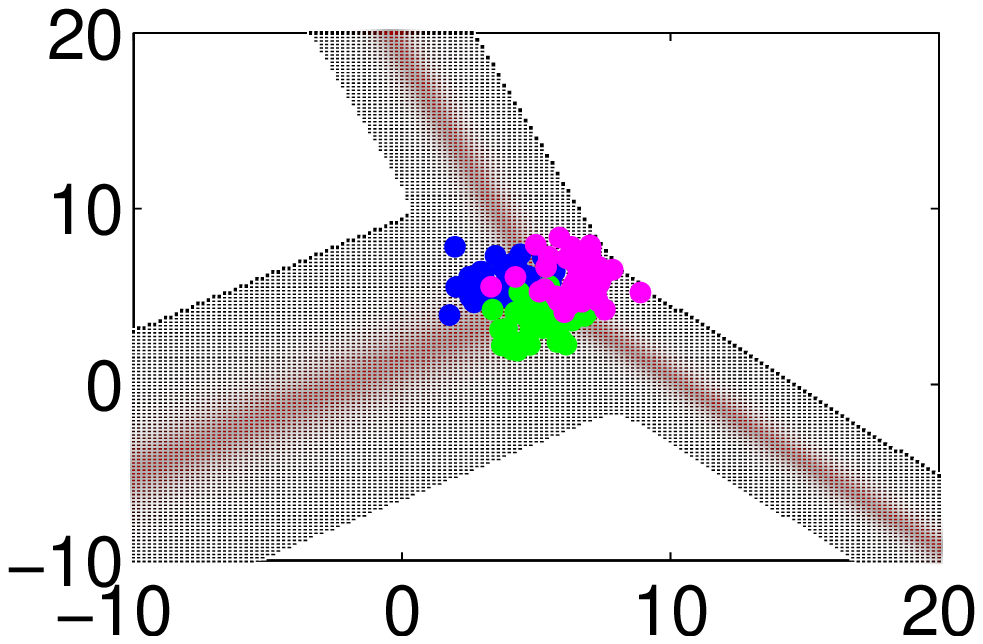} \\
    \multicolumn{2}{l}{\footnotesize{~~(a) Data1: Soft Const. ($88.33\%$) \qquad \quad ~ (b) Data1: Hard Const. ($83.33\%$)}} \\
    \\
    \includegraphics[width=0.235\textwidth]{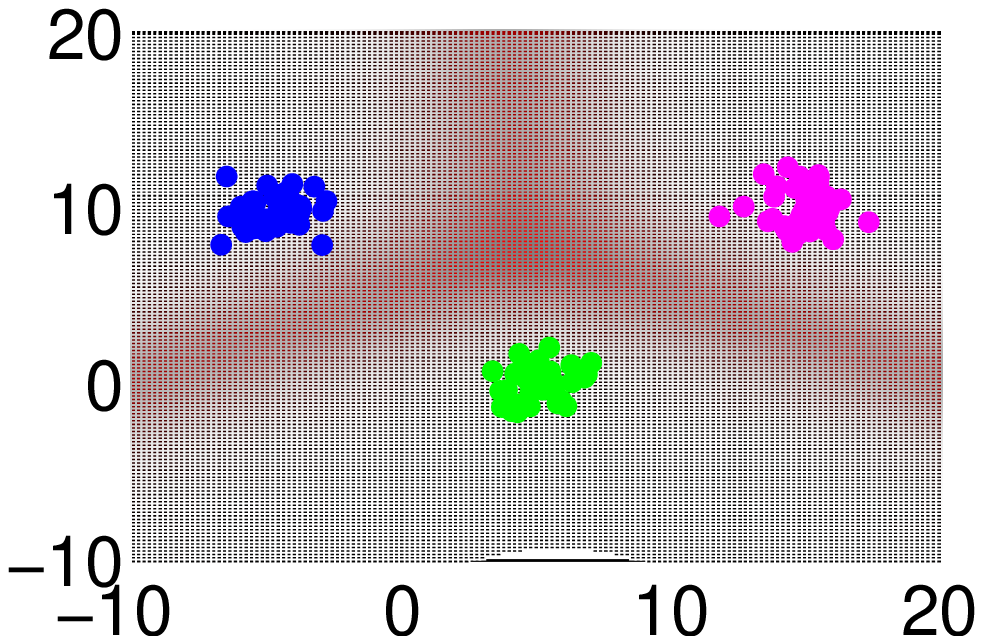}
   & \hspace{-0.1cm}
    \includegraphics[width=0.235\textwidth]{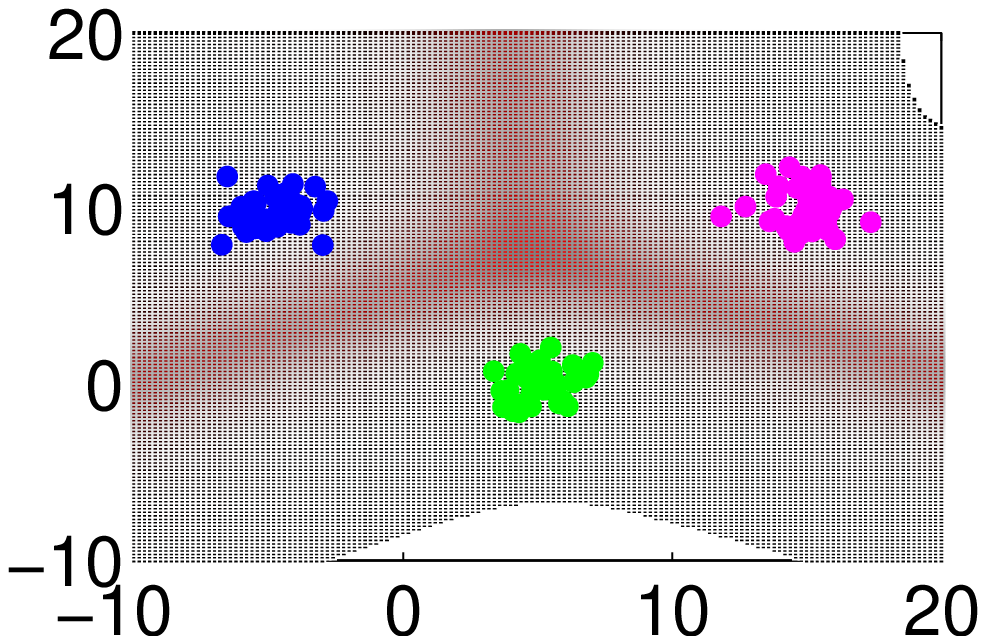} \\
    \multicolumn{2}{l}{\footnotesize{~~(c) Data2: Soft Const. ($100\%$)  \qquad \qquad ~ (d) Data2: Hard Const. ($100\%$)} }\\
\end{tabular}
\caption{\label{fig:entropy}  Estimated entropy using soft/hard constraints on synthetic datasets.
    Cluster assignments are represented with blue, pink, and green points.
    Entropy regions are shaded, with darker color representing higher entropy.
    Prediction accuracy on instance cluster labels is shown in the parentheses. }
\end{figure}%
In this set of experiments, we explore the impact on our model when soft constraints ($\epsilon=0.05$) and
hard constraints ($\epsilon=0$) are used respectively.
We first use two synthetic datasets to examine and illustrate their different behaviors.
These two datasets  each contain three clusters, 50 instances per cluster. The clusters are close to
each other in one dataset, and far apart (and thus easily separable) in the other.
For each dataset, we randomly generated 500 relative constraints using points near the decision boundaries.
Figure \ref{fig:entropy} shows the prediction entropy and prediction accuracy on instances cluster labels for both datasets
achieved by our model, using soft and hard constraints respectively. We can see that when
clusters are easily separable, both soft and
hard constraints produce reasonable decision boundaries and perfect prediction accuracy.
However, when cluster boundaries are fuzzy, the results of using soft constraints appear
preferable. This indicates that by \emph{softening} the constraints,
our method could search for more reasonable decision to avoid overfitting to the constrained instances.
%Thus we expect that using soft constraints would produce
%more promising results when the cluster boundaries are fuzzy.
%Our further experiments show that the difference becomes smaller as the constraints increase,
%and eventually they both lead to reasonably good decision boundaries.
%Due to lack of space, we do not present those results.
\begin{figure}
\centering
     \subfigure[Balance-scale]{ \includegraphics[width=0.237\textwidth]{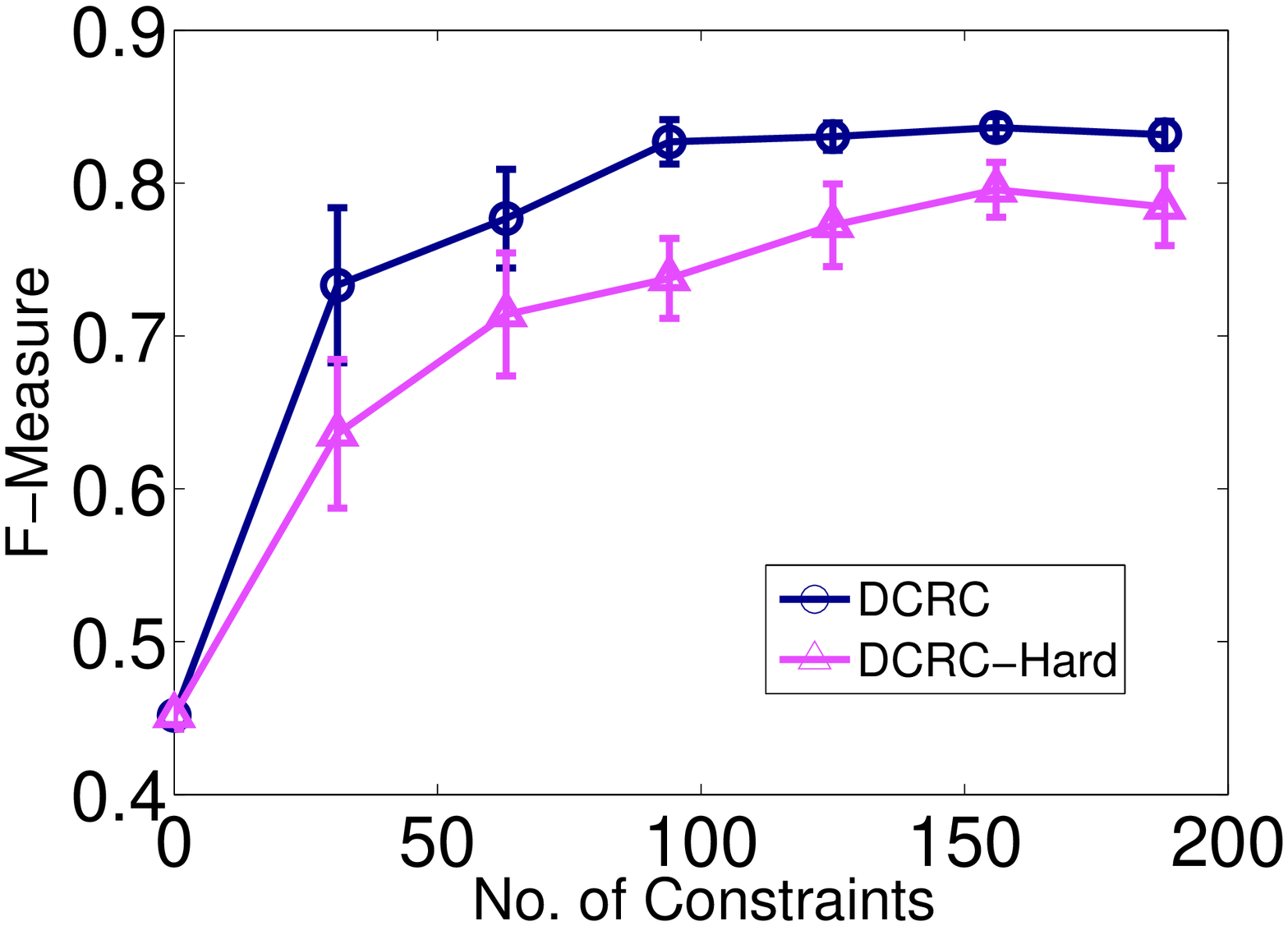} }
     \hspace{-0.6cm}
     \subfigure[Letters-IJLT]{ \includegraphics[width=0.237\textwidth]{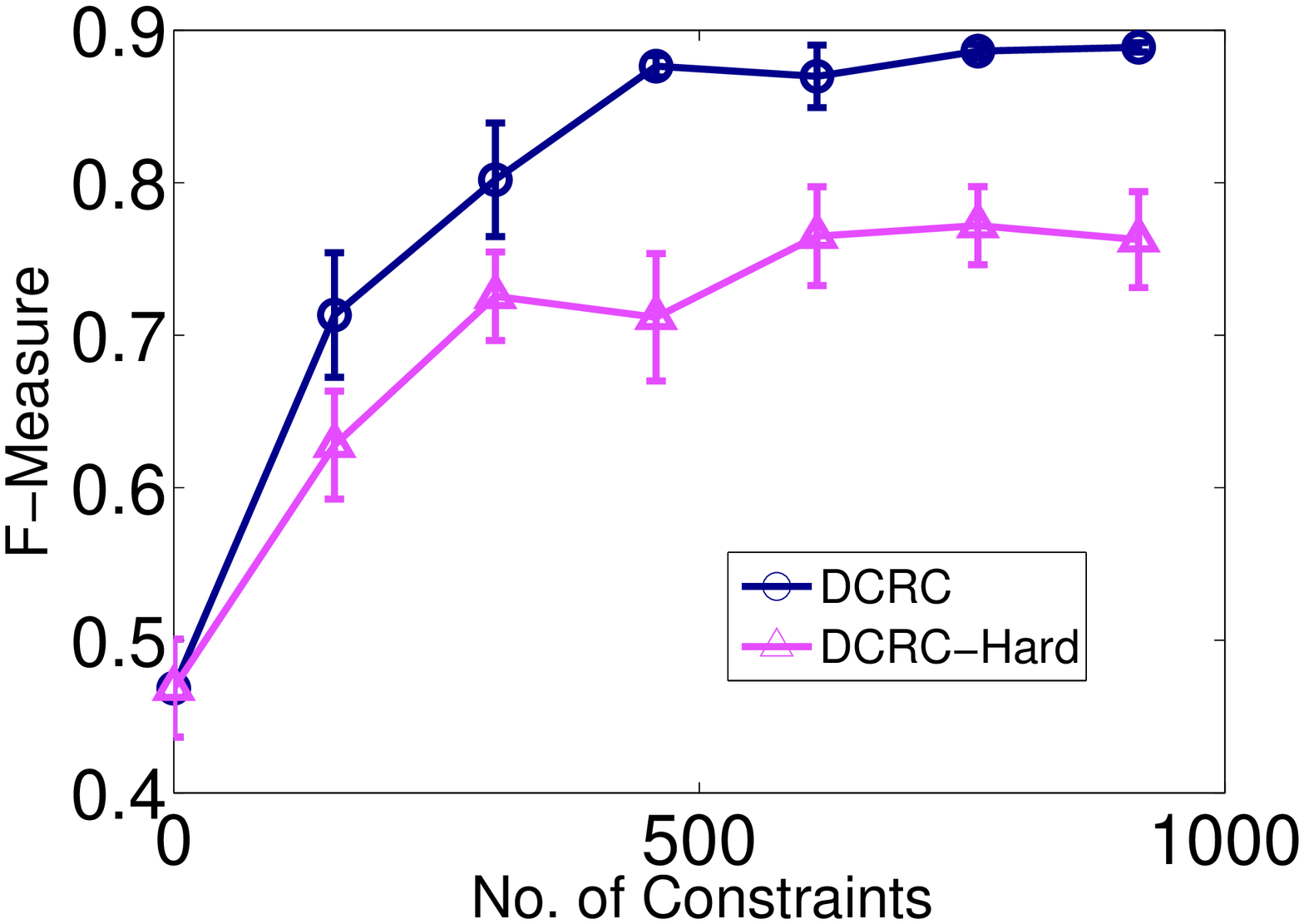} }
     \subfigure[MSRCv2]{ \includegraphics[width=0.237\textwidth]{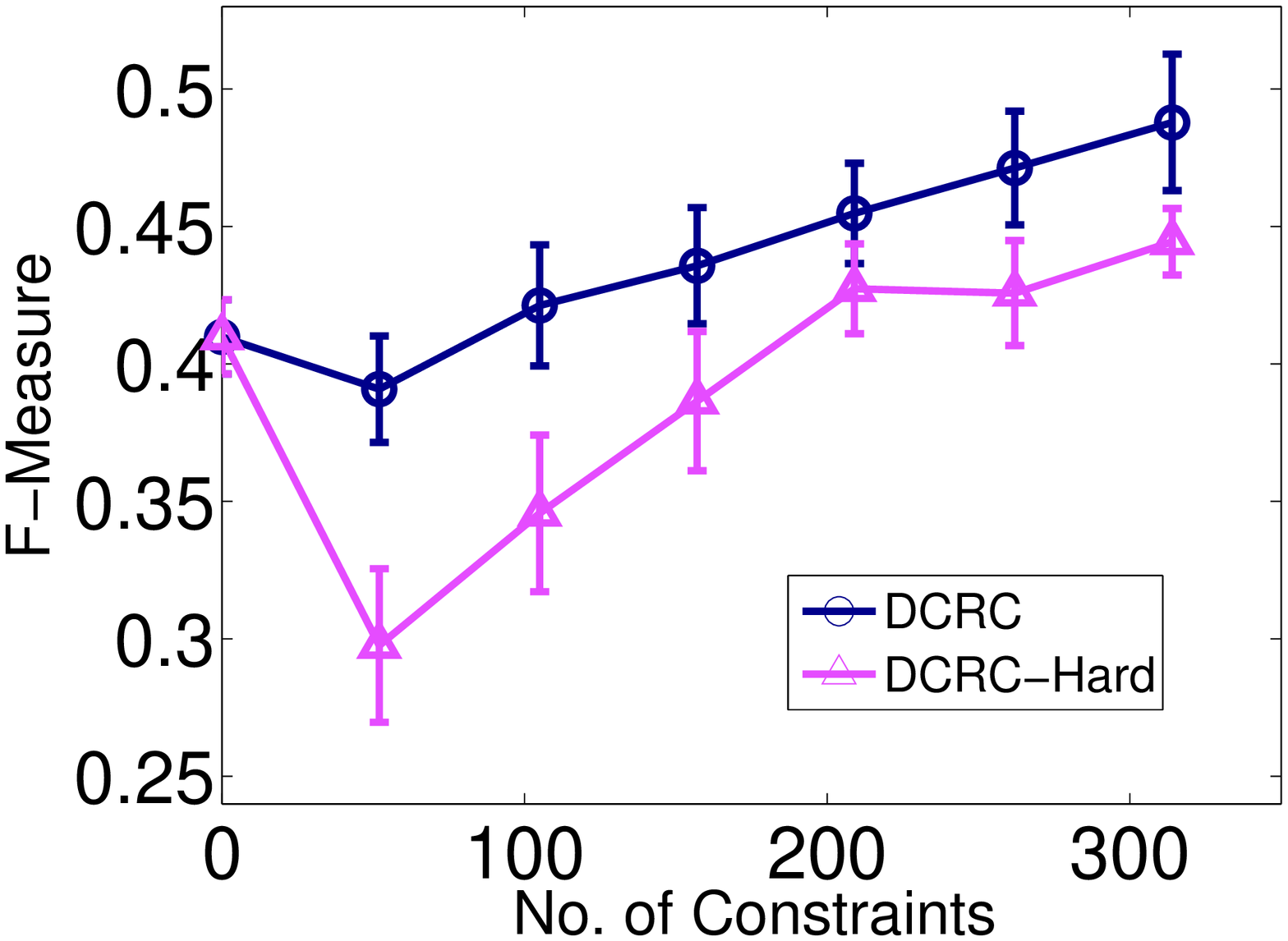} }
     \hspace{-0.6cm}
     \subfigure[Birdsong]{ \includegraphics[width=0.237\textwidth]{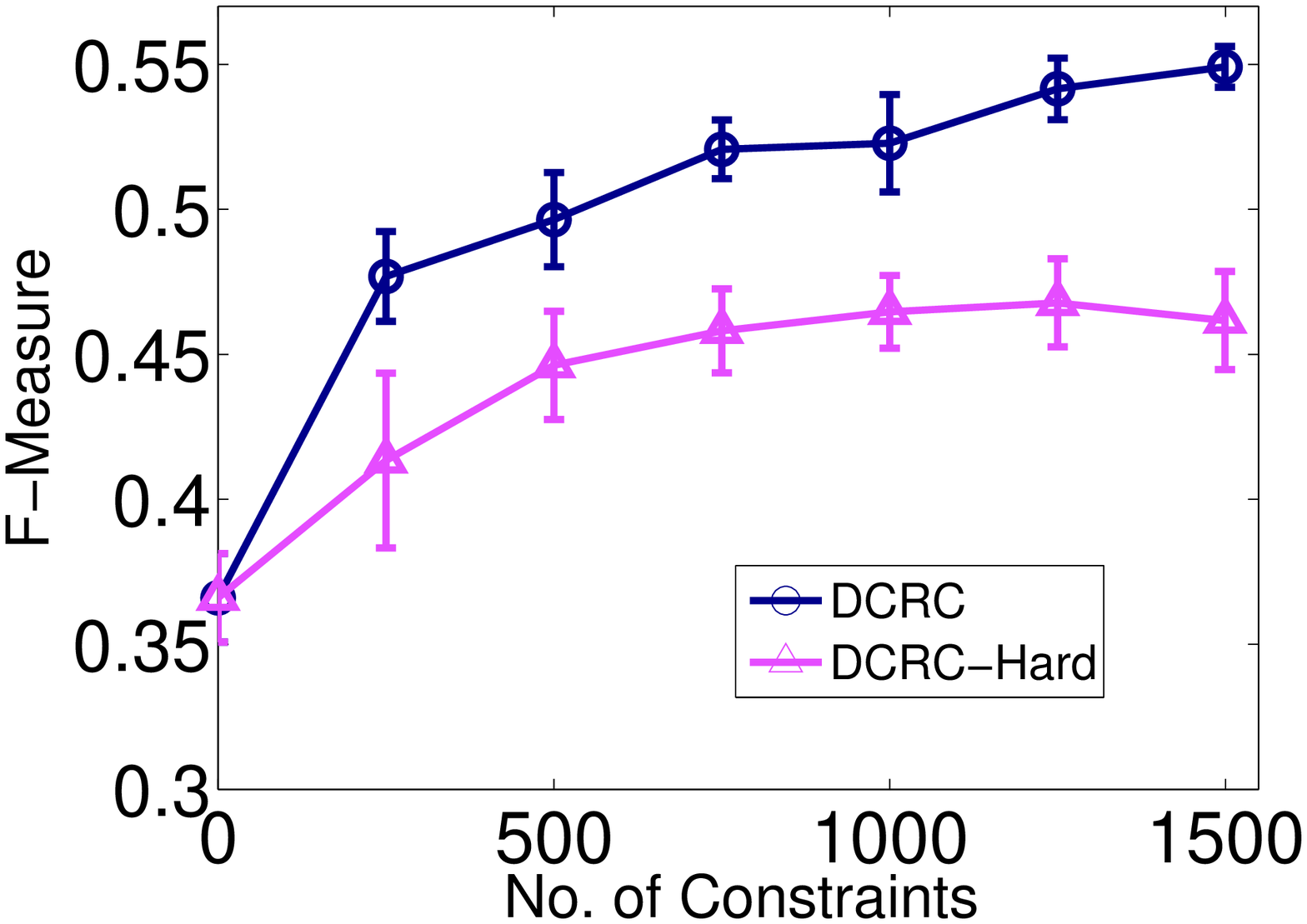} }
\caption{\label{fig:hardConst} Performance of DCRC using soft constraints vs.\ hard constraints. }
\end{figure}

We then compare the performances of using soft ($\epsilon=0.05$) versus
hard  ($\epsilon=0$) constraints on real datasets with the same setting
utilized in Section \ref{sec:const_size}. Due to space limit, here we only
show results on four representative datasets in Figure \ref{fig:hardConst}.
The behavior of other datasets are similar. %From Figure~\ref{fig:hardConst},
We can see that using soft constraints generally leads to better performance than
using hard constraints. In particular, on the \emph{MSRCv2} dataset, using hard constraints
produces a large ``dip'' at the beginning of the curve while this
issue is not severe for soft constraints. This suggests that using
soft constraints makes our model less susceptible to overfitting
to small sets of constraints.
%is especially preferred when the number of constraints is small.

\subsubsection{Effect of Cluster Balance Enforcement}
\begin{figure}
\centering
     \subfigure[Birdsong: Unbalanced]{ \includegraphics[width=0.237\textwidth]{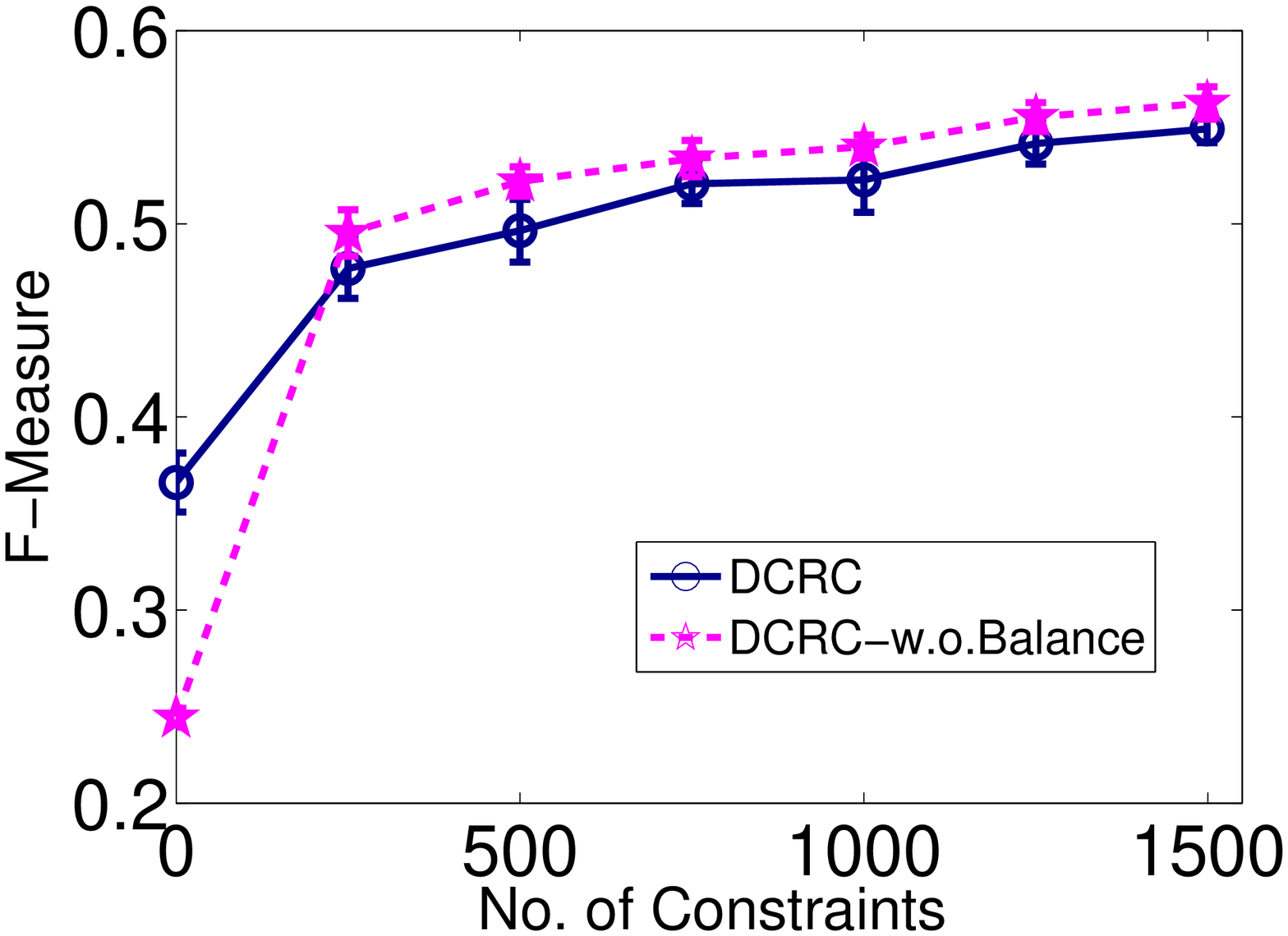} }
     \hspace{-0.6cm}
     \subfigure[Letters-IJLT: Balanced]{ \includegraphics[width=0.237\textwidth]{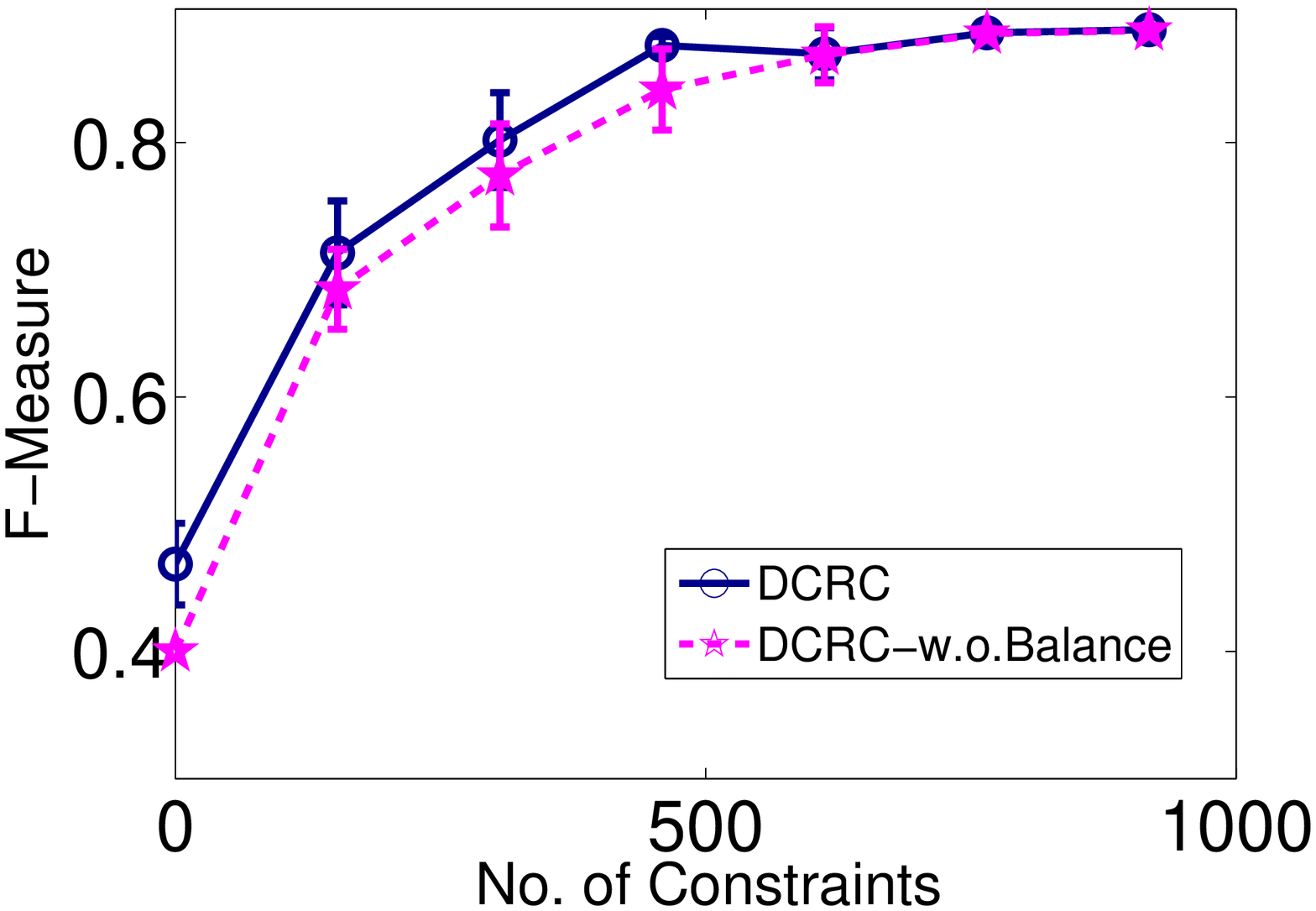} }
\caption{\label{fig:balance} Performance of DCRC with/without cluster balance enforcement.}
\end{figure}

%In our objective, cluster balance is enforced by imposing the maximization of the entropy of
%the estimated cluster label distribution.
This set of experiments test the effect of the cluster balance enforcement on the performance of DCRC for the
unbalanced \emph{Birdsong} and the balanced \emph{Letters-IJLT} datasets.
Figure \ref{fig:balance} reports the performance of DCRC (soft constraints, $\epsilon=0.05$) with and without
such enforcement with varied number of constraints.
We see that when there is no constraint, it is generally
beneficial to enforce the cluster balance. The reason is, when cluster balance is not enforced,
the entropy that enforces cluster separation can be trivially reduced
by removing cluster boundaries, causing degenerate solutions.
However, as the constraint increases, enforcing cluster balance on the unbalanced \emph{Birdsong}
hurts the performance. Conceivably, such enforcement would cause DCRC to prefer solutions
with balanced cluster distributions, which is undesirable for datasets with uneven classes.
On the other hand, appropriate enforcement on the balanced \emph{Letters-IJLT} dataset
provides further improvement. % over DCRC without cluster balance.
In practice, one could determine whether to enforce cluster balance
based on prior knowledge of the application domain.

\subsubsection{Computational Time}
We record the runtime of learning with 1500 constraints on the
Birdsong dataset, on a standard desktop computer with 3.4 GHz
CPU and 11.6 GB of memory. On average it takes less than 2 minutes to train the model
using an un-optimized Matlab implementation. This is reasonable for most applications
with similar scale.

\subsection{Case Study: Human-labeled Constraints}
We now present a case study where we investigate the impact of human-labeled constraints on the proposed method and its competitors.
%We also evaluate the performance of all the methods on these human-labeled constraints.

\subsubsection{Dataset and Setup}
This case study is situated in one of our applications where the
goal is to find bird singing patterns by clustering. The birdsong dataset
used in Section \ref{sec:noisefree} contains
%instances corresponding to automatically generated
spectrogram segments labeled with bird species. In reality,
birds of the same species may vocalize in different patterns, which we hope to
identify as different clusters. Toward this goal, we created another birdsong
dataset consisting of clusters that contains relatively pure singing patterns.
We briefly describe the data generation process as follows.

We first manually selected a collection of representative examples of the singing patterns, % we wish to capture,
and then use them as templates to extract segments from %a collection of
birdsong spectrograms by applying template matching. Each of the extracted segments
is assigned to the cluster represented by the corresponding template.
We then manually inspected and edited the clusters to ensure the quality of the clusters.
As a result, each cluster contains relatively pure segments
that are actually from the same bird species and represent the same vocalization pattern.
See Figure~\ref{fig:example} for examples of several different vocalization patterns,
which we refer to as syllables. We extract features for each segment using the
same method as described in \cite{briggs12KDD}. This process results in a
new Birdsong dataset containing $2601$ instances and $14$ ground-truth clusters.
%\footnote{We have made the data segments and features available at \url{http://web.engr.oregonstate.edu/~peiy/user_study/}}.

After obtaining informed consents according to the protocol approved by the Institutional
Review Board of our institution, % Oregon State University,
we tested six human subjects' behaviors on labeling constraints.
None of the users has any prior experience/knowledge on the data.
They were first given a short tutorial on the data and the concepts of \emph{clustering} and \emph{constraints}.
Then each user is asked to label  randomly selected  $150$ triplets, % (with \emph{yes}/\emph{no}/\emph{dnk}),
and $225$ pairs, % (with \emph{CL}/\emph{ML}),
using a graphical interface that displays the spectrogram segments.
To neutralize the potential bias introduced by the task ordering (triplets vs.\ pairs),
we randomly split the users into two groups with each group using a different ordering.

\begin{table}
\caption{\label{tab:crosstab}The average confusion matrix of the human labeled constraints  vs.\
    the constraint labels inferred from true instance clusters.}
\centering
\subtable[Relative Constraints]{ \label{tab:crosstab_a}
    \begin{tabular}{|l|ccc|}
    \hline
         \multirow{2}{*}{True} &   \multicolumn{3}{c|}{Human Labels } \\
                 &  $yes$  &  $no$ & $dnk$\\ \hline
              $yes$  &  18.50 & 0.33  &4.83  \\
              $no$ &   0.33 & 16.50 &4.50  \\
              $dnk$ &   3.50 & 3.00  &98.50 \\
    \hline
     \end{tabular}
     }
     \hspace{-0.3cm}
\subtable[Pairwise Constraints]{
    \renewcommand*{\arraystretch}{1.1}
   \begin{tabular}{|l|cc|}
    \hline
         \multirow{2}{*}{True} &   \multicolumn{2}{c|}{Human Labels } \\
                               &  $ML$  &  $CL$\\ \hline
                         $ML$  &  42.50  &  9.67   \\
                         $CL$  &   10.83 & 162.00  \\
    \hline
     \end{tabular}
     }
\end{table}
\subsubsection{Results and Discussion}
Table \ref{tab:crosstab} lists the average confusion matrix of the
human-labeled constraints versus the labels produced based on the ground-truth cluster labels.
From Table \ref{tab:crosstab_a}, we see that the \emph{dnk}
constraints make up more than half of the relative constraints, which is consistent with
our analysis in Section \ref{sec:info} that the number of \emph{dnk} constraints
can be dominantly large. The users rarely confuse between the \emph{yes} and \emph{no}
labels but they do tend to provide more erroneous \emph{dnk} labels. This phenomenon is
not surprising because when in doubt, we are often more comfortable to abstain from
giving an definite \emph{yes}/\emph{no} answer and resort to the \emph{dnk} option.

%This is conceivable since when three instances are far from
%their cluster centers, it is difficult for the users to provide
%\emph{yes}/\emph{no} answers with high confidence; and thus they tend
%to prefer a \emph{dnk} answer.
For pairwise constraints, the \emph{CL} constraints are the majority,
and the confusions for both \emph{CL} and \emph{ML} are similar.
We note that the confusion between the \emph{yes}/\emph{no} constraints
is much smaller than that of \emph{ML}/\emph{CL} constraints.
This shows that the increased flexibility introduced by \emph{dnk} label allows
the users to more accurately differentiate \emph{yes}/\emph{no} labels.
The overall labeling accuracy of pairwise constraints is slightly higher than that of relative
constraints. We suspect that this is due to the presence of the large amount of \emph{dnk} constraints.
%We expect that the labeling accuracy for relative constraints would become higher as
%\emph{yes}/\emph{no} constraints increase.
%
%\begin{table}
%\caption{\label{tab:userstudy}F-Measure performance (Mean $\pm$ Std) with the human labeled constraints. }
%\centering
% \subtable[Without using constraints]{\label{tab:userstudy_a}
%    \begin{tabular}{|c|c|}
%    \hline
%      Method & F-Measure              \\ \hline \hline
%     DCRC-NoConst   &   0.5175 $\pm$ 0.0232     \\
%     Kmeans&  0.6523 $\pm$ 0.0189  \\ \hline
%     \end{tabular}
%    }
% \subtable[Using 150 relative constraints]{
%    \begin{tabular}{|c|c|}
%    \hline
%    Method &  F-Measure  \\ \hline \hline
%        DCRC         &     0.7620 $\pm$ 0.1335  \\ %  \hline
%        DCRC-YN      &     0.7635 $\pm$ 0.1067  \\
%        LSML         &     0.6409 $\pm$ 0.0654  \\ %  \hline
%        sparseLP     &     0.5200 $\pm$ 0.0706  \\ %  \hline
%        SSSVaD       &     0.6046 $\pm$ 0.0605  \\  \hline
%     \end{tabular}
% }
% \subtable[Using pairwise constraints]{
%%    \label{tab:pair_150}
%    \begin{tabular}{|c|c|c|}
%      \hline
%   \multirow{2}{*}{Method} & \multicolumn{2}{c|}{F-Measure} \\ \cline{2-3}
%                &  150 constraints & 225  constraints \\ \hline \hline
%        ITML    &  0.6409 $\pm$ 0.0424 & 0.6347 $\pm$ 0.0372 \\ % \hline
% %       MPCKmeans      &  0.5617 $\pm$ 0.0236 & 0.5131 $\pm$ 0.0641 \\ % \hline
%        Xing    &  0.6438 $\pm$ 0.0423 & 0.6438 $\pm$ 0.0282 \\ \hline
%    \end{tabular}
%  }
%\end{table}
%
\begin{table}
\caption{\label{tab:userstudy}F-Measure performance (Mean $\pm$ Std) with the human labeled constraints. }
\centering
 \subtable[Without using constraints]{\label{tab:userstudy_a}
    \begin{tabular}{|c|c|}
    \hline
      Method & F-Measure              \\ \hline \hline
     DCRC-NoConst   &   0.5175 $\pm$ 0.0232     \\
     Kmeans&  0.6523 $\pm$ 0.0189  \\ \hline
     \end{tabular}
    }
 \subtable[Using 150 relative constraints]{
    \begin{tabular}{|c|c|}
    \hline
    Method &  F-Measure  \\ \hline \hline
        DCRC         &     0.7620 $\pm$ 0.1335  \\ %  \hline
        DCRC-YN      &     0.7635 $\pm$ 0.1067  \\
        LSML         &     0.6409 $\pm$ 0.0654  \\ %  \hline
        sparseLP     &     0.5200 $\pm$ 0.0706  \\ %  \hline
        SSSVaD       &     0.6046 $\pm$ 0.0605  \\  \hline
     \end{tabular}
 }
 \subtable[Using 150 pairwise constraints]{
    \begin{tabular}{|c|c|}
      \hline
       ~~~ Method~~~ & F-Measure \\ \hline \hline
        ITML    &  0.6409 $\pm$ 0.0424 \\ % \hline
        Xing    &  0.6438 $\pm$ 0.0423 \\ \hline
    \end{tabular}
  }
  %\hspace{0.2cm}
 \subtable[Using 225 pairwise constraints]{
    \begin{tabular}{|c|c|}
      \hline
       ~~~ Method ~~~ & F-Measure \\ \hline \hline
        ITML    &   0.6347 $\pm$ 0.0372 \\ % \hline
        Xing    &   0.6438 $\pm$ 0.0282 \\ \hline
    \end{tabular}
  }
\end{table}

We evaluated all the methods using these human-labeled
constraints. To account for the labeling noise in the constraints, we set
$\epsilon =0.15$ for DCRC and DCRC-YN\footnote{For these noisy constraints, our method remains
robust to the choice of $\epsilon$. Using different values of $\epsilon$ ranging
from $0.05$ to $0.2$ only introduces minor fluctuations (within 0.01 difference) to the F-measure.}.
The averaged results for all methods are listed in Table \ref{tab:userstudy}.
We observe that while most of the competing methods' performance degrade with
added constraints compared with unsupervised Kmeans, our method still shows
significant performance improvement even with the noisy constraints. We want to point out
that the performance difference we observe is not due to the use of the multi-class logistic classifier.
In particular, as shown in Table \ref{tab:userstudy_a}, without considering any constraints,
the logistic model achieves significantly lower performance than Kmeans.
This further demonstrates the effectiveness of our method in utilizing the side information
provided by noisy constraints to improve clustering.

Recall that  ITML is competitive with DCRC-YN previously considering noise-free constraints.
%It is interesting to note that
Here with noisy constraints, DCRC-YN achieves far better accuracy than ITML,
suggesting that our method is much more robust to labeling noise.
It is also worth noting that although the \emph{dnk} constraints tend to be quite noisy,
they do not seem to degrade the performance of DCRC compared with DCRC-YN.

Our case study also points to possible ways to further improve our model.
As revealed by Table \ref{tab:crosstab}, the noise on the labels for relative constraints is
not uniform as assumed by our model. An interesting future direction is to
introduce a non-uniform noise process to more realistically model the users' labeling behaviors.

\section{Related Work}
\label{sec:related}
{\it Clustering with Constraints:} Various techniques have been proposed for clustering
with pairwise constraints \cite{shental03NIPS,basu04KDD,lange05CVPR, lu04NIPS, nelson07ICML, lu07JMLR}.
Our work is aligned with most of these methods in the sense that we assume the
guidance for labeling constraints is the underlying instance clusters.

Fewer work has been done on clustering with relative constraints.
%Another branch of constrained clustering considers relative constraints
%(or relative comparisons) \cite{schultz2003NIPS,rosales2006KDD,kumar08TKDE,huang2011, liu2011KDD, liu2012ICDM}.
%Relative constraints encode instance similarity information by specifying answers to
%questions: is $x_i$ more similar to $x_j$ than to $x_k$.
The work in \cite{schultz2003NIPS,rosales2006KDD,kumar08TKDE,huang2011, liu2012ICDM} propose metric learning approaches
that use $d(x_i, x_j) < d(x_i, x_k)$ to encode that $x_i$ is more similar to $x_j$ than to $x_k$, where $d(\cdot)$ is the distance function.
%In addition to learning a metric, the method in \cite{kumar08TKDE} also finds a clustering solution  at the meantime.
The work \cite{liu2012ICDM} studies learning from relative comparisons between two pairs of instances,
which can be viewed as the same type of constraints when only three distinct examples are involved. % in the comparison.
By construction, these methods only consider constraints with \emph{yes}/\emph{no} labels.
Practically, such answers might not always be provided, causing limitation
of their applications. In contrast, our method is more flexible by allowing users to provide \emph{dnk} constraints,

There also exist studies that encode the  instance relative  similarities in the form of hieratical ordering
and attempt hierarchical algorithms that directly find clustering solutions satisfying the constraints  \cite{liu2011KDD,bade2013ML}.
Different with those studies, our work builds on a natural probabilistic model that has not been considered for learning with relative constraints.

{\it Semi-supervised Learning:} Related work also exists in a much broader area of semi-supervised learning, involving
studies on both clustering and classification problems. The work \cite{yves05NIPS} proposes that to
enforce the formed clusters with large separation margins, we could minimize the entropy on the unlabeled data,
in addition to learning from the labeled ones. The study \cite{gomes10NIPS} suggests to
also maximize the entropy of the cluster label distribution in order to find balanced clustering solution.
Our final formulation draws inspiration from the above work.

\section{Conclusions}
\label{sec:conclusion}
In this paper, we studied clustering with relative
constraints, where each constraint is generated by posing a query: {\it is
$x_i$ more similar to $x_j$ than to $x_k$}.  Unlike existing methods
that only consider \emph{yes}/\emph{no} responses to such
queries, we studied the case where the answer could also
be \emph{dnk} (don't know).
%, and formally showed how it is possible to obtain information from such responses.
%We also characterized how this setting compares with pairwise constraints based on
%the information content of the query responses.
We developed a probabilistic method DCRC that learns to cluster the instances
based on the responses acquired by such queries.
%We developed DCRC, a probabilistic method that learns to cluster
%from the responses provided by these queries.
% DCRC does not rely on learning a distance metric, and then using it to find a clustering. Instead, it
%employs the relationships between input instances, cluster
%labels, and query responses to find a probable clustering directly.
We empirically evaluated the proposed method using both simulated (noise-free) constraints
and human-labeled (noisy) constraints. The results demonstrated the usefulness of
\emph{dnk} constraints, the significantly improved performance of DCRC over existing
methods, and the superiority of our method in terms of the robustness to noisy constraints.

\bibliographystyle{IEEEtran}
\bibliography{triple_archive14}
%\begin{thebibliography}{1}
%\end{thebibliography}

\small
\appendix
This appendix provides the derivation of the mutual information
% between instance cluster labels and the relative constraint label, namely,
Eq.\ (3). The derivations for Eqns.\ (4) and (5)  are similar and are omitted here.

%\subsection{The Relative Constraint}
By definition, the mutual information between the instance labels $Y_t =[y_{t_1},y_{t_2},y_{t_3}]$
and the constraint label $l_t$  is
\begin{equation}
%\nonumber
 \label{eq_append:i_triple}
   I(Y_t;l_t) = \displaystyle  H(Y_t) - H(Y_t|l_{t}) .
\end{equation}
The first entropy term is
%\begin{equation}
%\nonumber
%\label{eq:hy_triple}
% \renewcommand{\arraystretch}{1.4}
%\begin{array}{l}
 $ H(Y_t)  = - \sum_{Y_t} P(Y_t) \log P(Y_t) = 3 \log K,$
% \end{array}
%\end{equation}
where we used the independence assumption
$P(Y_t) = \prod_{i=1}^3P(y_{t_i})$ and
substituted the prior $P(y_{t_i} = k) = {1}/{K}$.
By definition, the second entropy term is
%The conditional entropy of instance labels given the constraint label is
%\begin{equation}
%\label{eq:hy_l}
%\nonumber
% \renewcommand{\arraystretch}{1.3}
%\begin{array}{rl}
% \hspace{-0.2cm}  &  H(Y_t | l_t) \\
%                 =& \hspace{-0.3cm}{\sum \limits_{a \in \{\text{\emph{yes,no,dnk}} \} }} \hspace{-0.2cm} P(l_t=a) H(Y_t | l_t=a)\\
% \hspace{-0.2cm} =& -\hspace{-0.5cm}{\sum \limits_{a \in \{\text{\emph{yes,no,dnk}}\} }} \hspace{-0.2cm} P(l_t=a) {\sum \limits_{Y_t}} P(Y_t| l_t=a) \log P(Y_t| l_t=a) .
%\end{array}
%\end{equation}
\begin{equation}
\label{eq:hy_l}
\nonumber
      H(Y_t | l_t)
  =  -\hspace{-0.4cm}{\sum \limits_{a \in \{\text{\emph{yes,no,dnk}}\} }} \hspace{-0.4cm} P(l_t=a) {\sum \limits_{Y_t}} P(Y_t| l_t=a) \log P(Y_t| l_t=a).
\end{equation}
Now we need to compute the marginal distribution $P(l_t)$
and the conditional distribution $P(Y_t | l_t)$.
Based on  Eq.\ (1), the $P(l_t)$ are
%\vspace{-0.1cm}
\begin{equation}
 \nonumber
\label{eq:l_yes}
 \renewcommand{\arraystretch}{1.4}
\begin{array}{rl}
   P(l_t = \text{\emph{yes}})
    = &\hspace{-0.2cm} \sum_{Y_t} P(Y_t) P(l_t = \text{\emph{yes}}|Y_t) \\
 =&\hspace{-0.2cm} {\sum \limits_{k=1}^K} P(y_{t_1} = k) P(y_{t_2} = k) \left[ 1 - P(y_{t_3} = k) \right]
 =  \frac{K-1}{K^2}.
\end{array}
\end{equation}
By distribution symmetry,
 $P(l_t = \text{\emph{no}}) = P(l_t = \text{\emph{yes}})$. % = \frac{(K-1)}{K^2}
%\begin{equation}
% \nonumber
%\label{eq:l_no}
% \renewcommand{\arraystretch}{1.2}
%\begin{array}{rl}
%  P(l_t = \text{\emph{no}}) = P(l_t = \text{\emph{yes}}) = \frac{K-1}{K^2} ~.
%\end{array}
%\end{equation}
Then
$P(l_t = \text{\emph{dnk}})  =  1-  P(l_t = \text{\emph{yes}}) - P(l_t = \text{\emph{no}})
                             =  1 -  {[2(K-1)]}/{K^2}$.
%\begin{equation}
% \nonumber
%\label{eq:l_dk}
% \renewcommand{\arraystretch}{1.2}
%\begin{array}{l}
%   P(l_t = \text{\emph{dnk}})  =  1-  P(l_t = \text{\emph{yes}}) - P(l_t = \text{\emph{no}})
%                               =  1 - \frac{2(K-1)}{K^2} ~.
%\end{array}
%\end{equation}
To compute $P(Y_t | l_t)$, we notice that for the cluster label assignments that do not satisfy the
conditions for the corresponding $l_t$ described in Eq.\ (1), %\ref{tab:triple},
the probability $P(Y_t|l_t)=0$. For those satisfying such conditions, the
$P(Y_t|l_t)$ are
\begin{equation}
 \renewcommand{\arraystretch}{1.4}
 \nonumber
\begin{array}{rl}
  P(Y_t | l_t = \text{\emph{yes}})
  =&\hspace{-0.2cm} [P(Y_t)P(l_t = \text{\emph{yes}}|Y_t)] /P(l_t = \text{\emph{yes}}) \\
  =&\hspace{-0.2cm} [P(Y_t) \times 1]/P(l_t = \text{\emph{yes}})
  =  \frac{1}{K(K-1)} ~.
\end{array}
\end{equation}
%
%\begin{equation}
% \renewcommand{\arraystretch}{1.3}
% \nonumber
%\begin{array}{rl}
% \label{eq:yl_yes}
%  P(Y_t | l_t = \text{\emph{yes}})
%  =&\hspace{-0.2cm} [P(Y_t)P(l_t = \text{\emph{yes}}|Y_t)] /P(l_t = \text{\emph{yes}}) \\
%  =&\hspace{-0.2cm} [P(y_{t_1}) P(y_{t_2})P(y_{t_3}) \times 1]/P(l_t = \text{\emph{yes}}) \\
%  =&\hspace{-0.2cm} \frac{1}{K(K-1)} ~.
%\end{array}
%\end{equation}
By symmetry again,
$  P(Y_t | l_t = \text{\emph{no }}) =  P(Y_t|l_t = \text{\emph{yes}})$. % = \frac{1}{K(K-1)}
%\begin{equation}
% \nonumber
% \label{eq:yl_no}
% \renewcommand{\arraystretch}{1.4}
%    \begin{array}{ll}
%   P(Y_t | l_t = \text{\emph{no }}) = & P(Y_t|l_t = \text{\emph{yes}})
%                                    =  \frac{1}{K(K-1)} ~,
%\end{array}
%\end{equation}
Also,
\begin{equation}
 \renewcommand{\arraystretch}{1.4}
 \nonumber
\begin{array}{rl}
\label{eq:yl_dk}
  P(Y_t | l_t = \text{\emph{dnk}})
  =& \hspace{-0.2cm} [P(Y_t)P(l_t = \text{\emph{dnk}}|Y_t)] /P(l_t = \text{\emph{dnk}}) \\
  =& \hspace{-0.2cm} [P(Y_t) \times 1]/P(l_t = \text{\emph{dnk}})
  = \frac{1}{K[K^2-2(K-1)]} ~.
\end{array}
\end{equation}
Substituting the values of $P(Y_t|l_t)$ and $P(Y_t)$, we obtain
\begin{equation}
\nonumber
  H(Y_t | l_t) = \log K + (1- P_{\text{\emph{dnk}}}) \log (K-1)
                 + P_{\text{\emph{dnk}}} \log [K^2-2(K-1)],
\end{equation}
%\begin{equation}
%\nonumber
%\label{eq:hy_l_result}
% \renewcommand{\arraystretch}{1.2}
% \nonumber
%\begin{array}{rl}
%  H(Y_t | l_t) = &\log K + (1- P_{\text{\emph{dnk}}}) \log (K-1)\\
%                &\qquad\quad + P_{\text{\emph{dnk}}} \log [K^2-2(K-1)] ~,
%\end{array}
%\end{equation}
where we denote $P_{\text{\emph{dnk}}} = P(l_t = \text{\emph{dnk}})$.

Substituting $H(Y_t)$ and $H(Y_t|l_t)$
%Eq. (\ref{eq:hy_triple}) and Eq. (\ref{eq:hy_l_result})
into Eq.\ (\ref{eq_append:i_triple}), we derive
\begin{equation}
\nonumber
   I(Y_t;l_t)
 =  2\log K - (1-P_{\text{\emph{dnk}}})\log (K-1)
   - P_{\text{\emph{dnk}}} \log [K^2-2(K-1)]. ~~ \Box
\end{equation}
\end{document}